\documentclass[journal]{IEEEtran}

\usepackage[utf8]{inputenc} 
\usepackage[T1]{fontenc}    
\usepackage{hyperref}       
\usepackage{booktabs}       
\usepackage{amsfonts}       
\usepackage{nicefrac}       
\usepackage{microtype}      
\usepackage[pdftex]{graphicx}
\usepackage{amssymb,amsfonts,amsmath,amscd}
\usepackage[printonlyused,withpage]{acronym}
\usepackage{mathrsfs}
\usepackage{xcolor}
\usepackage{soul}
\usepackage{dsfont}
\usepackage{arydshln}
\usepackage{dblfloatfix}
\usepackage{afterpage}

\graphicspath{{figs/}}

\newcommand{\bbm}{\begin{bmatrix}}
\newcommand{\ebm}{\end{bmatrix}}
\newcommand{\mbf}{\mathbf}
\newcommand{\mbs}[1]{{\boldsymbol{#1}}}

\def\om{\omega}

\newcommand{\beq}{\begin{equation}}
\newcommand{\eeq}{\end{equation}}
\newcommand{\bdis}{\begin{displaymath}}
\newcommand{\edis}{\end{displaymath}}
\newcommand{\beqn}[1]{\begin{subequations}\label{eq:#1}\begin{eqnarray}}
\newcommand{\eeqn}{\end{eqnarray}\end{subequations}}
\newcommand{\est}[1]{\hat{#1}}
\newcommand{\pri}[1]{\check{#1}}
\newcommand{\wdg}{\wedge}

\newcommand{\Wdg}{\curlywedge}

\acrodef{BA}{Bundle Adjustment}
\acrodef{DNN}{Deep Neural Network}
\acrodef{EM}{Expectation Maximization}
\acrodef{GEM}{Generalized Expectation Maximization}
\acrodef{HMM}{Hidden Markov Model}
\acrodef{LDS}{Linear Dynamical System}
\acrodef{LQG}{Linear Quadratic Gaussian}
\acrodef{LQR}{Linear Quadratic Regulator}
\acrodef{LTI}{Linear Time-Invariant}
\acrodef{RTS}{Rauch-Tung-Striebel}
\acrodef{SGD}{Stochastic Gradient Descent}
\acrodef{SLAM}{Simultaneous Localization and Mapping}
\acrodef{RKHS}{Reproducing Kernel Hilbert Space}
\acrodef{SMW}{Sherman-Morrison-Woodbury}

\acrodef{GVI}{Gaussian Variational Inference}
\acrodef{ESGVI}{Exactly Sparse Gaussian Variational Inference}
\acrodef{MAP}{Maximum A Posteriori}
\acrodef{ML}{Maximum Likelihood}
\acrodef{KL}{Kullback-Leibler}
\acrodef{PDF}{Probability Density Function}
\acrodef{NEES}{Normalized Estimation Squared Error}
\acrodef{KF}{Kalman Filter}
\acrodef{VKF}{Variational Kalman Filter}
\acrodef{ISPKF}{Iterated Sigmapoint Kalman Filter}
\acrodef{ESGVI-GN}{ESGVI Gauss-Newton}
\acrodef{ELBO}{Evidence Lower Bound}
\acrodef{NGD}{Natural Gradient Descent}
\acrodef{FIM}{Fisher Information Matrix}
\acrodef{RANSAC}{Random Sample And Consensus}
\acrodef{IRLS}{Iteratively Reweighted Least-Squares}
\acrodef{BRD}{Black-Rangarajan Duality}
\acrodef{GNC}{Graduated Non-Convexity}
\acrodef{GP}{Gaussian process}

\hypersetup{%
    pdftitle={},
    pdfauthor={},
    pdfkeywords={},
    pdfsubject={},
    pdfstartview=FitH,%
    bookmarks=true,%
    bookmarksopen=true,%
    breaklinks=true,%
    colorlinks=true,%
    linkcolor=black,
    anchorcolor=black,%
    citecolor=black,
    filecolor=black,%
    menucolor=black,
    pagecolor=black,%
    urlcolor=black
}%

\graphicspath{{figs/}}

\begin{document}

\title{State Estimation for Continuum Multi-Robot Systems on SE(3)}

\author{Sven~Lilge, Timothy~D.~Barfoot, and~Jessica~Burgner-Kahrs
	\thanks{The authors are with the University of Toronto Robotics Institute, University of Toronto, Toronto, Ontario, Canada. e-mail: sven.lilge@utoronto.ca.}
	}

\markboth{December, 2024}%
{Lilge \MakeLowercase{\textit{et al.}}: State Estimation for Continuum Multi-Robot Systems on SE(3)}

\maketitle

\begin{abstract}
In contrast to conventional robots, accurately modeling the kinematics and statics of continuum robots is challenging due to partially unknown material properties, parasitic effects, or unknown forces acting on the continuous body.
Consequentially, state estimation approaches that utilize additional sensor information to predict the shape of continuum robots have garnered significant interest.
This paper presents a novel approach to state estimation for systems with multiple coupled continuum robots, which allows estimating the shape and strain variables of multiple continuum robots in an arbitrary coupled topology.
Simulations and experiments demonstrate the capabilities and versatility of the proposed method, while achieving accurate and continuous estimates for the state of such systems, resulting in
average end-effector errors of 3.3 mm and 5.02$^\circ$ depending on the sensor setup. 
It is further shown, that the approach offers fast computation times of below 10 ms, enabling its utilization in quasi-static real-time scenarios with average update rates of 100-200 Hz.
An open-source C++ implementation of the proposed state estimation method is made publicly available to the community.

\end{abstract}

\section{Introduction}

Continuum robots are flexible, slender manipulators largely inspired by the animal kingdom, resembling snakes, tentacles, elephant trunks, or worms \cite{Robinson1999,Burgner-Kahrs2015}.
Their elastic and jointless structure allows them to adhere to non-linear bending shapes, while being highly miniaturizable.
Owing to these properties, continuum robots can navigate in highly cluttered or confined spaces that traditional rigid-link robots typically cannot access.
This opens up several potential application areas with examples ranging from minimally invasive surgery \cite{Burgner-Kahrs2015} to industrial in-situ inspection and maintenance \cite{Dong2017,Wang2021} or search and rescue operations in disaster areas \cite{Hawkes2017}.

A recent trend in continuum robotics, inspired by parallel robots, involves arranging multiple coupled continuously deforming bodies in parallel assemblies to take advantage of the typical properties of parallel mechanisms, such as their increased stiffness and precision \cite{Bryson2014}.
The resulting structures are often refereed to as parallel continuum robots with early work focusing on designs comprising several passively deforming continuum links coupled to  a common end-effector platform.
Prominent examples include a continuous Stewart-Gough platform \cite{Black2018}, a continuum Delta robot \cite{Yang2018}, as well as mechanisms for planar positioning and orientating \cite{Mauze2020,Altuzarra2019b}, often presenting counterparts to existing parallel mechanisms with rigid links.
Instead of relying on passive continuum links, other previous work focuses on parallel assemblies of continuum bodies whose deformations are actively controlled, utilizing pneumatic pressure chambers \cite{Rivera2014,Lindenroth2019}, shape memory alloys \cite{Moghadam2015} or tendon actuation \cite{Nuelle2020,Lilge2022a}.
Related work envisions systems consisting of multiple collaborative continuum robots that can join together during operation to form coupled, parallel assemblies, adjusting their kinematic properties to benefit from an increased stiffness or precision \cite{Russo2022,Jalali2022}.
To date, research on both parallel and collaborative continuum robots is mostly concerned with their kinematic, static and dynamic modeling \cite{Lilge2020,Bryson2014,Till2019}, and the characterization of their properties, such as manipulability and compliance \cite{Black2018,Lilge2022a}, reachable workspace \cite{Zaccaria2022}, singularity conditions \cite{Briot2021,Lilge2022b} and stability \cite{Till2017}.
Additionally, novel designs are investigated, with examples ranging from reconfigurable mechanisms \cite{Mahoney2016} to structures featuring additional constraints \cite{Orekhov2017,Wu2019,Chen2019}.

\begin{figure}[t]
	\centering
	\scriptsize
	\def\svgwidth{1\linewidth}
\begingroup%
  \makeatletter%
  \providecommand\color[2][]{%
    \errmessage{(Inkscape) Color is used for the text in Inkscape, but the package 'color.sty' is not loaded}%
    \renewcommand\color[2][]{}%
  }%
  \providecommand\transparent[1]{%
    \errmessage{(Inkscape) Transparency is used (non-zero) for the text in Inkscape, but the package 'transparent.sty' is not loaded}%
    \renewcommand\transparent[1]{}%
  }%
  \providecommand\rotatebox[2]{#2}%
  \newcommand*\fsize{\dimexpr\f@size pt\relax}%
  \newcommand*\lineheight[1]{\fontsize{\fsize}{#1\fsize}\selectfont}%
  \ifx\svgwidth\undefined%
    \setlength{\unitlength}{990.95970334bp}%
    \ifx\svgscale\undefined%
      \relax%
    \else%
      \setlength{\unitlength}{\unitlength * \real{\svgscale}}%
    \fi%
  \else%
    \setlength{\unitlength}{\svgwidth}%
  \fi%
  \global\let\svgwidth\undefined%
  \global\let\svgscale\undefined%
  \makeatother%
  \begin{picture}(1,0.68042803)%
    \lineheight{1}%
    \setlength\tabcolsep{0pt}%
    \put(0,0){\includegraphics[width=\unitlength,page=1]{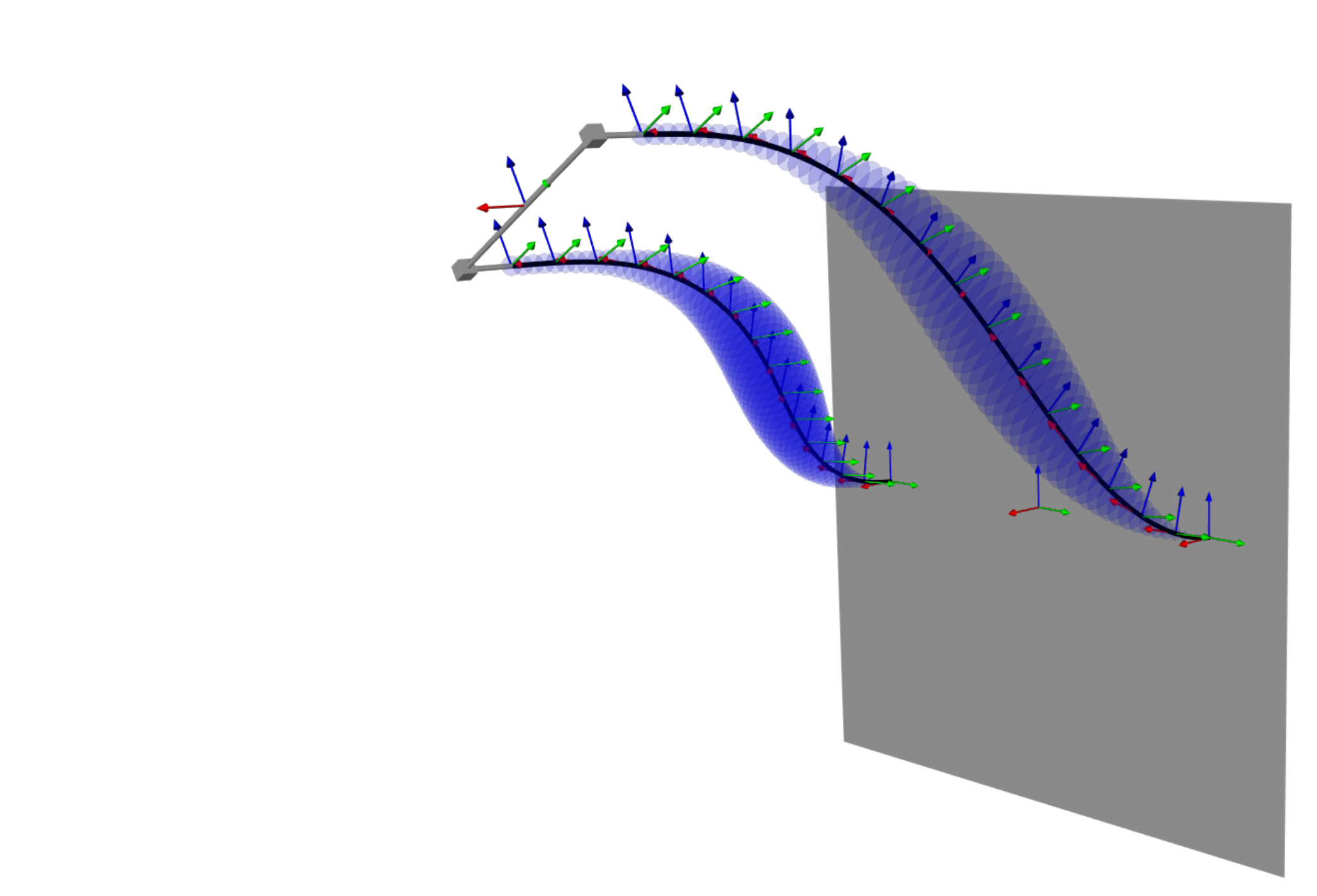}}%
    \put(0.83319428,0.64561605){\color[rgb]{0,0,0}\makebox(0,0)[t]{\lineheight{1.25}\smash{\begin{tabular}[t]{c}Estimated Robot \\ State Mean\end{tabular}}}}%
    \put(0.36099278,0.226079){\color[rgb]{0,0,0}\makebox(0,0)[t]{\lineheight{1.25}\smash{\begin{tabular}[t]{c}Uncertainty Ellipsoids\end{tabular}}}}%
    \put(0.27532567,0.64540527){\color[rgb]{0,0,0}\makebox(0,0)[t]{\lineheight{1.25}\smash{\begin{tabular}[t]{c}Common End-Effector \\ Pose Measurement\end{tabular}}}}%
    \put(0,0){\includegraphics[width=\unitlength,page=2]{figure1_example.pdf}}%
    \put(0.14405524,0.38360631){\color[rgb]{0,0,0}\makebox(0,0)[t]{\lineheight{1.25}\smash{\begin{tabular}[t]{c}Rigid Connections\end{tabular}}}}%
    \put(0,0){\includegraphics[width=\unitlength,page=3]{figure1_example.pdf}}%
  \end{picture}%
\endgroup%

	\caption{Example state estimate of a continuum multi-robot system consisting of two individual manipulators coupled to a common end-effector.}
	\label{fig:figure1}
\end{figure}

For control, accurate determination of the current state, such as shape or strain variables, of such systems is crucial.
While state-of-the-art modeling approaches for parallel and collaborative continuum robots achieve encouraging results, they exhibit non-negligible remaining errors.
Such errors usually arise from unmodeled, parasitic effects, and uncertainties in the assumed material properties of the manipulators.
Moreover, model-based methods require knowledge about the forces and moments acting on the elastic bodies, information that is usually not available when the robot is subject to complex interactions with its environment.
To compensate for model inaccuracies or unknown loading configurations, state estimation approaches using sensor information to reason about a continuum robot's state show promise.
Stochastic state estimation approaches are of particular interest, as they provide the probabilistic distribution of possible robot states given a prior model and sensor measurements.
This allows to not only obtain the robot state but also its likelihood, considering sensor noise and modeling uncertainties alike.

\subsection{Related Work}

In the following, we discuss related work on the state estimation for conventional and coupled continuum robots.
While several methods to infer the shape of continuum robots from sensor information exist, such as fitting certain kinematic representations \cite{Roesthuis2014,Kim2014,Song2015} or mechanics models \cite{Rone2013,Venkiteswaran2019} to sensor data, we are focusing our review on the most relevant stochastic state estimation approaches. 

Current state-of-the-art methods usually differ in their choice of a suitable underlying continuum robot model as well as their general approach to stochastic state estimation.
Generally, state estimation approaches might not need to employ the most sophisticated and complex mechanical model of continuum robots as they rely on additional sensor information.
Thus, the choice of the underlying model often revolves around choosing a suitable kinematic representation that can account for a variety of different shapes and states.
One common choice are constant-curvature models, which represent the kinematics of a continuum robot by a series of concatenated circular arcs \cite{Webster2010}.
State estimation approaches utilizing such models can, for instance, be found in \cite{Brij2010} and \cite{Borgstadt2015}, which both propose particle filter based approaches to estimate the tip pose or position of catheters over time.
In \cite{Ataka2016}, the state of a three-segment tendon-driven continuum robot is estimated as a sequence of tendon actuation vectors over time using a constant-curvature model in combination with an extended Kalman filter.
A similar approach for multi-backbone continuum robots is presented in \cite{Chen2019a}, where an unscented Kalman filter is used to track the arc parameters of the robots' constant-curvature representation over time.
Lastly, \cite{Loo2019} present a state estimation approach for a soft robot modelled as a circular arc, using a Kalman filter to track a sequence of bending curvature angles over time.

While the assumption of constant-curvature deformations for continuum robots may be reasonable for certain types of robots operating in free space, the accuracy of such kinematic representations is limited, especially in the presence of external forces and moments.
As an alternative, \cite{Lobaton2013} proposes a continuum robot state estimation approach using learned shape basis functions as a kinematic representation, being able to account for more general shapes.
A Kalman filter is used to estimate the coefficients of these learned basis functions over time.

State estimation approaches with more general kinematic representation can be found in \cite{Mahoney2016a} and \cite{Anderson2017}.
Both employ Kirchhoff rod models to represent continuum robots, allowing to estimate their states, including curvature, position and orientation, as general continuous functions over their length, instead of relying on a finite number of approximating parameters.
The estimation approach in both works utilizes a Rauch-Tung-Striebel smoother to solve for the full robot state along their length.
In contrast to the previously discussed works, which aimed to track the states of continuum robots over time, the state estimation problem here is formulated with respect to the length of the continuum robot, resulting from the nature of the employed Kirchhoff rod equations. 
The proposed method is applied to both a concentric tube continuum robot \cite{Mahoney2016a} as well as a reconfigurable parallel continuum robot \cite{Anderson2017} by taking into account coupling constraints between individual manipulators.
At runtime, the differential Kirchhoff rod equations are integrated numerically using a shooting method approach, while considering boundary conditions and coupling constrains.

Lastly, we presented a general state estimation approach applicable to any continuum robot that can be modeled with Cosserat rod theory \cite{Lilge2022}.
Here, the state of continuum robots is represented using Gaussian processes along their lengths, allowing to describe them as continuous functions with uncertainty.
The state estimation approach then makes use of Gaussian process regression to incorporate noisy sensor measurements and to compute the whole posterior of the continuous state as a batch problem.

In conclusion, while a variety of different approaches exist for the state estimation of continuum robots, almost all approaches exclusively handle conventional serial continuum manipulators.
To the best of our knowledge, the work of \cite{Anderson2017} currently presents the only state estimation approach that allows to explicitly consider multiple coupled continuum structures.

\subsection{Contributions}

In this article, we are extending our prior work on the state estimation of continuum robots using Gaussian process regression on $SE(3)$ \cite{Lilge2022} to systems that consist of multiple coupled continuum manipulators.
In order to do so, each continuum robot is described as an individual Gaussian process and additional cost terms are incorporated into a sparse factor graph representation to consider coupling between the individual robots.
This formulation results in a general framework that is able to handle any topology of coupled continuum robots in a straightforward manner (see Fig.~\ref{fig:figure1} for an example).
We believe it will be particularly useful to estimate the state of parallel continuum robots \cite{Black2018,Lilge2022a} and collaborative continuum robots subject to coupling constraints \cite{Russo2022,Jalali2022}.

We are highlighting the capabilities of the proposed state estimation approach in simulations, where different robot topologies with varying coupling assemblies and sensor setups are considered.
To prove the generality of the approach, it is applied to several existing structures featuring multiple coupled continuum robots from the current state of the art.
Further, experiments on a robotic prototype are conducted for additional quantitative evaluations.
Here, we particularly show how our method can be used in conjunction with commonly employed sensor technologies for continuum robots, such as electromagnetic tracking coils and fiber Bragg grating sensors.
Suitable sensor models are derived for each of such sensors to be used in our state estimation framework.
Lastly, we show that our state estimation framework can be implemented in a highly efficient manner by exploiting the underlying sparse structure of the resulting factor graph representation.
Depending on the system's topology and the sensor availability, such an implementation is capable of computing the state estimation problems in real time, allowing its potential use in online scenarios such as real-time closed-loop control.
To facilitate the usage of our proposed method, an efficient C++ implementation of our approach is made openly available to the community.

We note that our proposed work is similar to the work of \cite{Anderson2017}, which presents a state estimation approach for coupled continuum robots for the first time.
However, we believe that there are several striking differences that set our work apart.
First, due to using a general prior based on a simplified Cosserat rod model our method can be applied to any continuum robot, regardless of its geometry or actuation principle.
Second, we show that our approach can consider any topology of continuum robots in a straightforward manner.
The shooting method approach \cite{Anderson2017} is only evaluated on one particular type of parallel continuum robot.
It might be non-trivial to adapt it to some of the more intricate coupling topologies presented in this manuscript and it further remains unclear how well the method would converge in these cases.
Third, we show that our method is highly efficient, offering computational rates that allow usage in online scenarios.
On the contrary, the runtime of the shooting-based approach in \cite{Anderson2017} is not evaluated.
While numerical integration schemes can potentially be carried out quickly, we believe that it likely runs slower than the method presented here.
We derive closed-form analytical expressions for the stochastic integrals, which are efficient to compute at runtime.
Lastly, stating our state estimation approach as a general optimization problem allows to potentially consider additional constraints and cost terms.
This could be beneficial for a number of use cases, such as dealing with contact constraints in a known environment, for which our approach could easily be extended in the future.

\begin{figure}[b]
	\centering
	\footnotesize
	\def\svgwidth{0.95\linewidth}
\begingroup%
  \makeatletter%
  \providecommand\color[2][]{%
    \errmessage{(Inkscape) Color is used for the text in Inkscape, but the package 'color.sty' is not loaded}%
    \renewcommand\color[2][]{}%
  }%
  \providecommand\transparent[1]{%
    \errmessage{(Inkscape) Transparency is used (non-zero) for the text in Inkscape, but the package 'transparent.sty' is not loaded}%
    \renewcommand\transparent[1]{}%
  }%
  \providecommand\rotatebox[2]{#2}%
  \newcommand*\fsize{\dimexpr\f@size pt\relax}%
  \newcommand*\lineheight[1]{\fontsize{\fsize}{#1\fsize}\selectfont}%
  \ifx\svgwidth\undefined%
    \setlength{\unitlength}{189.71987266bp}%
    \ifx\svgscale\undefined%
      \relax%
    \else%
      \setlength{\unitlength}{\unitlength * \real{\svgscale}}%
    \fi%
  \else%
    \setlength{\unitlength}{\svgwidth}%
  \fi%
  \global\let\svgwidth\undefined%
  \global\let\svgscale\undefined%
  \makeatother%
  \begin{picture}(1,0.44097905)%
    \lineheight{1}%
    \setlength\tabcolsep{0pt}%
    \put(0,0){\includegraphics[width=\unitlength,page=1]{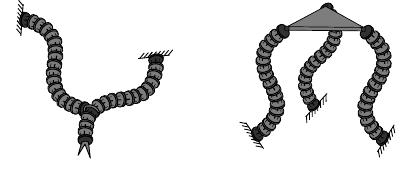}}%
    \put(0.47605507,0.11728774){\color[rgb]{0,0,0}\rotatebox{0.0000034}{\makebox(0,0)[t]{\lineheight{1.25}\smash{\begin{tabular}[t]{c}Coupling\end{tabular}}}}}%
    \put(0.56330828,0.40954177){\color[rgb]{0,0,0}\rotatebox{0.0000034}{\makebox(0,0)[t]{\lineheight{1.25}\smash{\begin{tabular}[t]{c}Common End-Effector\end{tabular}}}}}%
    \put(0.21172509,0.00037321){\color[rgb]{0,0,0}\rotatebox{0.0000034}{\makebox(0,0)[t]{\lineheight{1.25}\smash{\begin{tabular}[t]{c}Collaborative Continuum Robots\end{tabular}}}}}%
    \put(0.80108171,0.00121998){\color[rgb]{0,0,0}\rotatebox{0.0000034}{\makebox(0,0)[t]{\lineheight{1.25}\smash{\begin{tabular}[t]{c}Parallel Continuum Robot\end{tabular}}}}}%
    \put(0,0){\includegraphics[width=\unitlength,page=2]{multi_robot_example.pdf}}%
  \end{picture}%
\endgroup%

	\caption{Example continuum multi-robot systems considered for the proposed state estimation approach; Left: \textit{Collaborative continuum robots} subject to coupling; Right: \textit{Parallel continuum robot} consisting of multiple individual manipulators coupled to a common end-effector platform.}
	\label{fig:robot_architectures_se}
\end{figure}

\section{Continuum Multi-Robot Systems}

Throughout this work, we are considering \textit{continuum multi-robot systems}, which consist of $N$ individual continuum robots that may be subject to coupling constraints with respect to each other.
We additionally consider rigid objects, to which the continuum robots can be coupled.
One typical example for such a rigid object is a common end-effector platform to which the continuum robots are physically attached, which is usually the case for parallel continuum robots \cite{Black2018}.

The to-be-estimated state $\mathbf{x}$ of the resulting robot system includes the shape and strain of each continuum robot in addition to the pose of the common end-effector to which the robots are coupled.
Throughout the remainder of this chapter, we will use the terms \textit{system state}, referring to all of the quantities making up the whole state of the robot system, and \textit{continuum robot state}, referring to the shape and strain of an individual continuum robot.

The complete system state $\mathbf{x}$ consists of
\begin{align}
\mathbf{x}_n(s_n) &= \{ \mathbf{T}_n(s_n), \mbs{\varepsilon}_n(s_n) \}, \quad \text{with} \quad n \in \{1,...,N\}, \label{eq:state1} \\ 
\mathbf{x}_\text{ee} &= \mathbf{T}_\text{ee}. \label{eq:state2}
\end{align}
Here $\mathbf{x}_n(s_n)$ is the state of the $n$th continuum robot in a system of $N$ robots, consisting of its pose $\mathbf{T}_n(s_n) \in SE(3)$ and strain $\mbs{\varepsilon}_n(s_n)  \in \mathds{R}^6$ along its arclength $s_n$.
The common end-effector platform's state $\mathbf{x}_\text{ee}$ consists of its pose $\mathbf{T}_\text{ee} \in SE(3)$.

We note that, unless stated otherwise, all transformations throughout this paper are defined as a transformation $\mathbf{T}_{bi}$ from the inertial frame $\left\{i\right\}$, which is a static world frame, to the body frame $\left\{b\right\}$, attached to the continuum robot or end-effector platform.
It is further noted that the robot architectures considered in this chapter can consist of any number of these individual parts of the system.
For instance, designs can consist of only a single continuum robot, multiple coupled continuum robots without a common end-effector, or systems featuring both multiple robots and a common end-effector platform.
For completeness, the derivations throughout this chapter are all expressed considering systems with both multiple coupled continuum robots and a common end-effector, but are applicable to all possible robot architectures.
Lastly, while the examples discussed in this work only consider the existence of one common end-effector platform, the shown derivations can easily be adapted to handle any number of rigid bodies included in the coupled system.

Following this definition of continuum multi-robot systems, Fig.~\ref{fig:robot_architectures_se} shows some example architectures that the proposed state estimation approach can handle, including \textit{collaborative continuum robots} as well as \textit{parallel continuum robots}.
Note that these two simple examples are the most common types of coupled continuum robots that currently exist.
However, our proposed state estimation method is able to handle even more complex topologies and we will show some more intricate examples later throughout this paper.

\section{Continuum Robot Model Prior}

Following our prior work \cite{Lilge2022}, we represent each individual continuum robot with their own Gaussian process prior based on a simplified Cosserat rod model.
We do not assume any prior knowledge about the state of the common end-effector and its pose will later be inferred based on known coupling constraints and sensor readings.

In the following, we review and summarize the development of the employed continuum robot prior.
We refer the reader to \cite{Lilge2022} for a more detailed derivation of the corresponding terms.

\subsection{Simplified Cosserat Rod Model}

Each continuum robot in our multi-robot system is modeled using Cosserat rod theory, a widely used approach for modeling continuum robots.
While existing Cosserat rod model formulations generally depend on the continuum robot architecture, type and actuation principle \cite{Rucker2010,Rucker2011}, we will utilize a simplified, more general model.

We describe the continuous state of each continuum robot using the following set of differential equations according to Cosserat rod theory:
\begin{align}\label{eq:cosserat_model}
\frac{d}{ds_n}\mbf{T}_n(s_n) &= \mbs{\varepsilon}_n(s_n)^\wdg \mbf{T}_n(s_n), \\
\frac{d}{ds_n}  \mbs{\varepsilon}_n(s_n) &= \mbs{\mathcal{K}}_n^{-1}(\mbs{f}(s_n)-\mbs{\varepsilon}(s_n)^{\Wdg^T} \mbs{\sigma}_n(s_n)), \label{eq:ds_strain}
\end{align}
where $\mbs{\mathcal{K}}_n \in \mathds{R}^{6\times6}$ is the square stiffness matrix of the continuum robot, $\mbs{f}(s_n) \in \mathds{R}^6$ are distributed external forces and moments applied to its body and $\mbs{\sigma}_n(s_n)  \in \mathds{R}^6$ is its internal stress.
Further, $\mbs{\varepsilon}_n(s_n)^\wdg$ and $\mbs{\varepsilon}_n(s_n)^\Wdg$ are defined as
\begin{align}
\mbs{\varepsilon}_n(s_n)^\wdg &= \bbm \mbs{\nu}_n(s_n) \\ \mbs{\om}_n(s_n) \ebm^\wdg = \bbm \mbs{\om}_n(s_n)^\wdg & \mbs{\nu}_n(s_n) \\ \mbf{0}^T & 0 \ebm,
\\ \mbs{\varepsilon}_n(s_n)^\Wdg &= \bbm \mbs{\nu}_n(s_n) \\ \mbs{\om}_n(s_n) \ebm^\Wdg = \bbm \mbs{\om}_n(s_n)^\wdg & \mbs{\nu}_n(s_n)^\wdg  \\ \mbf{0} & \mbs{\om}_n(s_n)^\wdg \ebm,
\end{align}
where $\mbs{\om}_n(s_n) \in \mathds{R}^3$ and $\mbs{\nu}_n(s_n) \in \mathds{R}^3$ are the robot's rotational and translational strain variables and the $^\wdg$ operator maps a vector from $\mathds{R}^3$ to a skew-symmetric matrix in $\mathds{R}^{3\times3}$.
Throughout this work, elongation and twisting strains are defined in the $x$-axis, while bending and shearing strains are defined in the $y$- and $z$-axes.
This leads to continuous robot shapes whose length is defined along their $x$-axis.

\subsection{Gaussian Process Prior in $SE(3)$}

Our next goal is to construct a Gaussian process prior for each continuum robot in the form
\begin{equation}\label{eq:prior}
\mbf{x}_n(s_n) \sim \mathcal{GP}( \pri{\mbf{x}}_n(s_n), \pri{\mbf{P}}_n(s_n,s_n^\prime)), 
\end{equation}
where $\pri{\mbf{x}}_n(s_n)$ and $\pri{\mbf{P}}_n(s_n,s_n^\prime)$ are the prior mean and covariance functions, respectively.

First, in order to deal with the non-linearities of our differential equations, we use a series of local Gaussian processes that are stitched together.
For this, we discretize the continuous state of each continuum robot into $K_n$ discrete nodes at arclengths $s_{n,k}$.
We initialize a local Gaussian process at each of those arclengths and define local pose variables in the Lie algebra, $\mbs{\xi}_{n,k}(s_n) \in \mathfrak{se}(3)$.
With this, we can define the continuous pose between two discrete arclengths, $s_{n,k}$ and $s_{n,k+1}$, as
\begin{equation}\label{eq:localvar}
\mathbf{T}_n(s_n) = \underbrace{\exp\left( \mbs{\xi}_{n,k}(s_n)^\wdg \right)}_{\in \, SE(3)} \mathbf{T}_n(s_{n,k}).
\end{equation}

In the following, we now replace the second derivative of this local pose variable with a zero-mean, white-noise Gaussian process such that
\begin{equation}\label{eq:noise_second_dir}
	\frac{d^2}{ds^2_n}  \mbs{\xi}_{n,k}(s_n) = \mathbf{w}_{n,k}(s_n), \quad \mathbf{w}_{n,k}(s_n) \sim \mathcal{GP}( \mbf{0}, \mbf{Q}_c(s-s^\prime)).
\end{equation}
Here, $\mbf{Q}_c(s-s^\prime)$ is the covariance function of the Gaussian process and $\mbf{Q}_c$ is a stationary power-spectral density matrix, i.e., the continuous version of a covariance matrix.
This has several important implications.
First, this means that we assume no prior knowledge about the forces and moments acting on the continuum robot, either from external loads or from actuation, which might be available from more sophisticated physics-based models \cite{Black2018,Lilge2022a}.
While this might seem to be a fairly conservative standpoint, we assume that we will later have sufficient sensor and coupling information to accurately estimate the state of the robot.
On top of that, this assumption makes our state estimation approach applicable to any continuum robot structure without requiring knowledge about its type or actuation principle.
Second, by setting the second derivative of the local pose variable to a zero-mean, white-noise Gaussian process, our prior will favour robot configurations in which the derivative of the local strain is equal to zero.
The employed continuum robot prior can thus be interpreted as a \textit{constant-strain prior}, in which robot configurations with a constant strain, or near-constant strain, are more likely to occur.

We can now express our state equations using the local pose variable $\mbs{\xi}_{n,k}(s_n)$, which results in a first-order stochastic differential equation
\begin{equation}\label{eq:locallti}
\frac{d}{ds_n} \bbm \mbs{\xi}_{n,k}(s_n) \\  \mbs{\psi}_{n,k}(s_n) \ebm = \bbm \mathbf{0} & \mathbf{1} \\ \mathbf{0} & \mathbf{0} \ebm \underbrace{\bbm \mbs{\xi}_{n,k}(s_n) \\  \mbs{\psi}_{n,k}(s_n) \ebm}_{\mbs{\gamma}_{n,k}(s_n)} + \bbm \mathbf{0} \\ \mathbf{1} \ebm \mathbf{w}_{n,k}(s_n),
\end{equation}
where $\mbs{\gamma}_{n,k}(s_n)$ is the Markovian state, $\mbs{\psi}_{n,k}(s_n)~=~\frac{d}{ds_n}\mbs{\xi}_{n,k}(s_n)$, and $\mathbf{1}$ is the identity matrix.

Since using local variables results in a linear first-order differential equation, we can stochastically integrate \eqref{eq:locallti} in closed form to obtain
\begin{align}\label{eq:GP}
\mbs{\gamma}_{n,k}(s_n) &\sim \mathcal{GP} \bigl( \underbrace{\mbs{\Phi}(s_n,s_{n,k}) \pri{\mbs{\gamma}}_{n,k}(s_{n,k})}_{\rm mean~function}, \notag \\ &\underbrace{\mbs{\Phi}(s_n,s_{n,k}) \pri{\mathbf{P}}_n(s_{n,k})  \mbs{\Phi}(s_n,s_{n,k})^T + \mathbf{Q}(s_n-s_{n,k})}_{\rm covariance~function} \bigr).
\end{align}
Here, $\mbs{\Phi}(s,s^\prime)$ is the {\em transition function},
\begin{equation} 
\mbs{\Phi}(s,s^\prime) = \bbm \mathbf{1} & (s-s^\prime) \mathbf{1} \\ \mathbf{0} & \mathbf{1} \ebm, \quad s \geq s^\prime,
\end{equation}
$\mathbf{Q}(s-s^\prime)$ is the covariance accumulated between two arclengths,
\begin{equation}
\mathbf{Q}(s-s^\prime) = \bbm \frac{1}{3} (s-s^\prime)^3 \mathbf{Q}_c & \frac{1}{2} (s-s^\prime)^2 \mathbf{Q}_c \\ \frac{1}{2} (s-s^\prime)^2 \mathbf{Q}_c & (s-s^\prime) \mathbf{Q}_c \ebm, \quad s \geq s^\prime,
\end{equation}
and $\pri{\mbs{\gamma}}_{n,k}(s_{n,k})$ and $\pri{\mathbf{P}}_n(s_{n,k})$ are the initial mean and covariance at $s_n = s_{n,k}$, the starting point of the local variable.
$\mathbf{Q}_c$ is a stationary power-spectral density matrix, i.e., the continuous version of the covariance matrix, and its elements can be tuned to control the \textit{smoothness} and behavior of the continuum robot prior.

Later, we will further need to express our local state variables with respect to the global ones, which can be achieved using
\begin{align}
	\mbs{\xi}_{n,k}(s_n) &= \ln \left( \mbf{T}(s_n) \mbf{T}(s_{n,k})^{-1} \right)^\vee, \label{eq:local_global1}\\
	\mbs{\psi}_{n,k}(s_n) &= \frac{d}{ds_n}\mbs{\xi}_{n,k}(s) = \mbs{\mathcal{J}}\left( \mbs{\xi}_{n,k}(s_n) \right)^{-1} \mbs{\varepsilon}_n(s_n). \label{eq:local_global2}
\end{align}
where $\ln( \cdot)$ is the matrix logarithm, $\vee$ is the inverse operator of $\wdg$ and $\mbs{\mathcal{J}}$ is the left Jacobian of $SE(3)$ \cite[p.301]{Barfoot2024}.

\begin{figure*}[t]
	\centering
	\footnotesize
	\def\svgwidth{1\linewidth}
	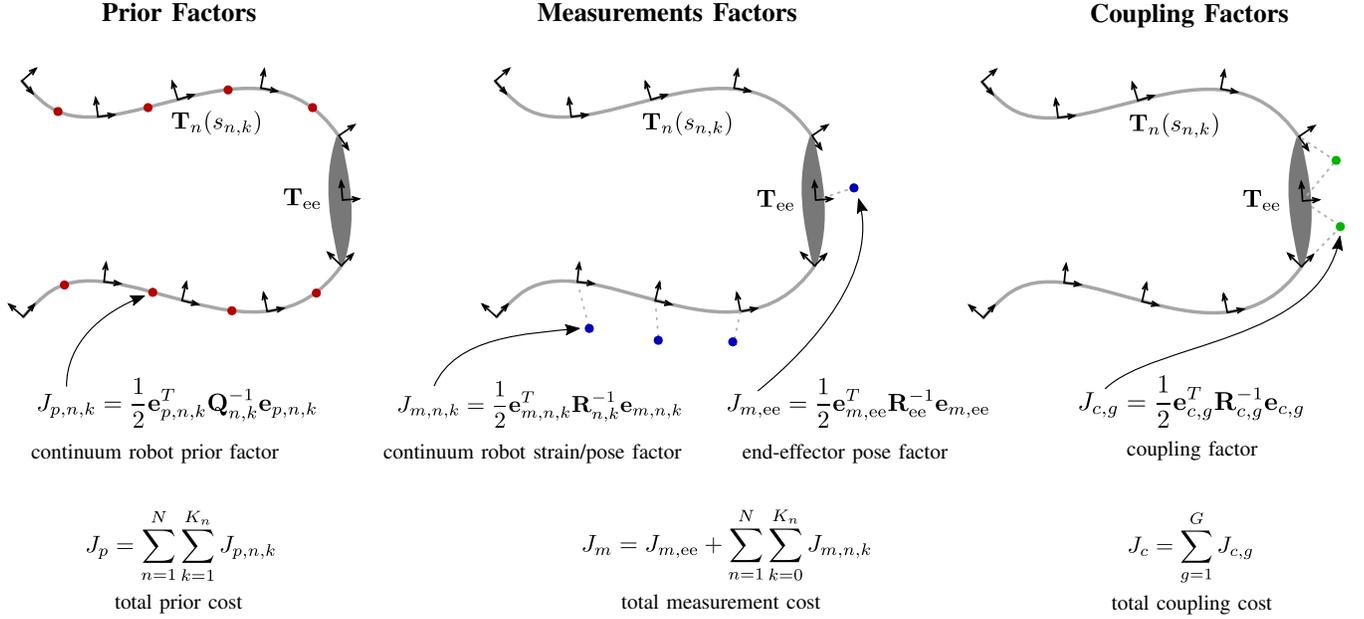
	\caption{Factor graph representation of an example multi-robot system, consisting of two continuum robots coupled to a common end-effector. Left: The prior cost terms for each continuum robot are represented by binary factors, each involving two consecutive discrete states (red dots). Middle: The measurement cost terms are represented by unary factors, each involving only one discrete state that is associated with the measurement (blue dots). Right: The coupling cost terms are represented by binary factors, each involving two discrete states that are subject to a coupling constraint with respect to each other (green dots).}
	\label{fig:factor_graph}
\end{figure*}

\section{Maximum A Posteriori Objective Function}

In the following, we formulate a batch state estimation for all variables in our system state according to \eqref{eq:state1} and \eqref{eq:state2} using a Maximum A Posteriori (MAP) objective.
This section discusses the construction of the overall objective function for the MAP approach, whose minimizing state will be the mode of the posterior, i.e., the most likely state of our system state. 
The objective function consists of three main terms, one for the prior expressions of the continuum robots, one for measurements, and one for coupling constraints between the individual parts of our system.
By representing our system as a \textit{factor graph}, each individual component in our objective function corresponds to one factor in the graph, involving one or several discrete states of our system.
Fig.~\ref{fig:factor_graph} visualizes the factor graph with factors for the prior, measurements, and coupling constraints for an example system consisting of two continuum robots coupled to a common end-effector platform.

\subsection{Prior Cost Terms}

Following \cite{Lilge2022}, we define the error according to the prior between two sequential discrete states of continuum robot $n$ as 
\begin{multline}\label{eq:localerr}
\mathbf{e}_{p,n,k} = \left( \mbs{\gamma}_{n,k}(s_{{n,k}}) - \pri{\mbs{\gamma}}_{n,k}(s_{n,k})\right) - \hspace{2.4cm} \\ \mbs{\Phi}(s_{n,k},s_{n,k-1}) \left( \mbs{\gamma}_{n,k}(s_{n,k-1}) - \pri{\mbs{\gamma}}_{n,k}(s_{n,k-1})\right).
\end{multline}
We can further construct the following squared-error cost term to represent the negative log-likehood of this error:
\begin{equation}\label{eq:cost}
J_{p,n,k} = \frac{1}{2} \mathbf{e}_{p,n,k}^T \mathbf{Q}_{n,k}^{-1} \mathbf{e}_{p,n,k},
\end{equation}
where $\mathbf{Q}_{n,k} = \mathbf{Q}(s_{n,k}-s_{n,k-1})$.
Each cost term is a \textit{binary factor} in our factor graph (see Fig.~\ref{fig:factor_graph}, left) and expresses how close the two corresponding consecutive states are to our constant-strain prior formulation.
Using \eqref{eq:local_global1} and \eqref{eq:local_global2} we can formulate the cost term using our original global variables as \cite{Anderson2015a}
\begin{equation}\label{eq:globalerr}
	\mathbf{e}_{p,n,k} = \bbm  \mbs{\xi}_{n,k,k-1} - (s_{n,k} - s_{n,k-1}) \, \mbs{\varepsilon}_n(s_{n,k-1}) \\ \mbs{\mathcal{J}}\left(  \mbs{\xi}_{n,k,k-1} \right)^{-1} \mbs{\varepsilon}_n(s_{n,k}) - \mbs{\varepsilon}_n(s_{n,k-1}) \ebm,
\end{equation}
where
\begin{equation}
	\mbs{\xi}_{n,k,k-1} = \ln \left( \mathbf{T}_n(s_{n,k}) \mathbf{T}_n(s_{n,k-1})^{-1} \right)^\vee.
\end{equation}
Considering all $K_n$ discrete states, the individual prior errors for each continuum robot $n$ can be summed up as
\begin{equation}
J_{p,n} = \sum_{k=1}^{K_n} J_{p,n,k}.
\end{equation}
The total prior cost of our system can be written as
\begin{equation}
J_{p} = \sum_{n=1}^{N} J_{p,n} = \sum_{n=1}^{N}\sum_{k=1}^{K_n} J_{p,n,k},
\end{equation}
summing up the prior cost terms for each continuum robot $n$.

\subsection{Measurement Cost Terms}

We will consider two different types of noisy measurements for our state estimation, one related to measuring discrete poses and one related to measuring the strain at discrete arclengths of the continuum robots.
In both cases, each measurement introduces a new cost term, which can be represented as a {\em unary factor} in the factor graph, involving the discrete state of our system associated with the respective measurement (see Fig.~\ref{fig:factor_graph}, middle).
The resulting cost terms for both types of measurements are discussed in the following.

\subsubsection{Pose Measurements}

For the first type of measurement, we assume that we can measure the full pose of particular discrete states in our system, including the poses of the individual continuum robots at discrete arclengths and the pose of the common end-effector.
This can, for instance be done using electromagnetic (EM) tracking coil sensors.

Following our prior work \cite{Lilge2022}, we can define a noisy pose measurement $\widetilde{\mathbf{T}}_{n,k}$ of continuum robot $n$ at discrete arclength $s_{n,k}$ as
\begin{equation}
	\widetilde{\mathbf{T}}_{n,k}  = \exp\left( \mathbf{n}_{n,k}^\wdg \right) \mathbf{T}_n(s_{n,k}),
\end{equation}
where $\mathbf{T}_n(s_{n,k})$ is the true pose, $\mathbf{n}_{n,k} \in \mathbb{R}^6$ is a regular Gaussian random variable drawn from $\mathcal{N}\left( \mathbf{0}, \mathbf{R}_{n,k} \right)$ and $\mathbf{R}_{n,k} \in \mathbb{R}^{6\times 6}$ is the covariance associated with the expected measurement noise.
Using this measurement equation, we can formulate the error for this pose measurement as
\begin{equation}\label{eq:poseerr}
\mathbf{e}_{m,n,k} = \ln \left(  \mathbf{T}_n(s_{n,k})\widetilde{\mathbf{T}}_{n,k}^{-1} \right)^\vee,
\end{equation}
which will be zero, if our measured state matches the estimated state.
The squared-error cost term, representing the negative log-likelihood of this error, can be written as
\begin{equation}\label{eq:posecost}
J_{m,n,k} = \frac{1}{2} \mathbf{e}_{m,n,k}^T \mathbf{R}_{n,k}^{-1} \mathbf{e}_{m,n,k}.
\end{equation}

Following the same logic, we can define analogous expressions for noisy pose measurements of the common end-effector.
The error term can be written as
\begin{equation}\label{eq:poseerr_ee}
\mathbf{e}_{m,\text{ee}} = \ln \left(  \mathbf{T}_\text{ee}\widetilde{\mathbf{T}}_\text{ee}^{-1} \right)^\vee,
\end{equation}
where $\mathbf{T}_\text{ee}$ is the estimated pose of the end-effector and $\widetilde{\mathbf{T}}_\text{ee}$ is its measured pose.
The squared-error cost term for the pose measurement of the end-effector is 
\begin{equation}\label{eq:posecost_ee}
J_{m,\text{ee}} = \frac{1}{2} \mathbf{e}_{m,\text{ee}}^T \mathbf{R}_{\text{ee}}^{-1} \mathbf{e}_{m,\text{ee}},
\end{equation}

For both the measurements of the pose at certain arclength of the continuum robots as well as of the end-effector pose, a projection matrix can be used to mask off unmeasured degrees of freedom in~\eqref{eq:poseerr} and~\eqref{eq:poseerr_ee}.
This can, for instance, be useful when using EM tracking coils that are unable to measure roll, as a twisting motion around their longitudinal axis has no effect on the measured current induced by the external electromagnetic field.
Another example is the incorporation of position measurements without information about the orientation, e.g., from cameras.

\subsubsection{Strain Measurements}
\label{sec:sensor_model}
In our prior work \cite{Lilge2022}, we explored incorporating sensor readings to measure the strain variables $\mbs{\varepsilon}_n(s_n)$ in continuum robots. 
While direct strain measurements are challenging to achieve, optical fibers with inscribed fiber Bragg gratings (FBG) provide an indirect approach \cite{Ryu2014}. 
These gratings reflect specific light wavelengths, shifting with longitudinal strain and temperature. 
By assuming constant temperature, gratings act as optical strain gauges, correlating wavelength shifts with longitudinal strain. 
Multiple gratings in a fiber enable strain measurements at various locations. 
Typically, these fibers are arranged in predefined patterns within a sensor, facilitating the reconstruction of the sensor array's shape by correlating longitudinal strains with curvature strains \cite{Moore2012}.

The literature offers various reconstruction models for continuum robot strain and shape sensing \cite{Ryu2014,Xu2016,Modes2020}, but their integration into stochastic state estimation frameworks remains underexplored. 
Our work addresses this gap, detailing the application of a specific FBG sensor model within a stochastic framework, enhancing our understanding of uncertainty in reconstruction techniques due to sensitivity and sensor noise.

\begin{figure}[t]
	\centering
	\includegraphics[width=0.6\linewidth]{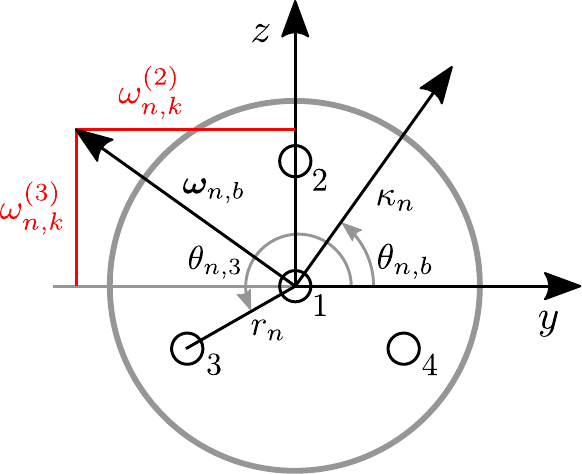}
	\caption{FBG sensor cross-section of continuum robot $n$ showing the parallel arrangement of four individual optical fibers. The outer fibers are arranged in a circular pattern with distance $r_n$ to the center and angles $\theta_{n,i}$ with respect to the body frame. The overall bending  occurs about the the vector $\boldsymbol{\omega}_{n,b}$, with curvature $\kappa_n$ and bending angle $\theta_{n,b}$.
	$\boldsymbol{\omega}_{n,b}$ is orthogonal to the bending plane and directly related to the bending curvature strains $\omega^{(2)}_{n,k}$ and $\omega^{(3)}_{n,k}$.}
	\label{fig:fbg_cross_section}
\end{figure}

We focus on multi-core FBG sensors comprising four optical fibers in parallel (Fig.~\ref{fig:fbg_cross_section}), with one central and three peripherally arranged fibers at distance $r_n$ and angles $\theta_{n,i}$. Initial models assumed bending deformations \cite{Moore2012}, but recent advancements include twisting and elongation deformations \cite{Modes2020}.
We extend these developments to construct a sensor model for our state estimation framework, leveraging the capabilities of FBG sensors.

Let us assume that we have noisy measurements $\widetilde{\mathbf{y}}_{n,k}~\in~\mathds{R}^4$ of the longitudinal strain values $\boldsymbol{\lambda}_{n,k}~\in~\mathds{R}^4$ for each grating $i~\in~\{1,...,4\}$ of the four fibers at discrete arclengths $s_{n,k}$ of a continuum robot $n$.
We can write the following expression
\begin{align}
	\widetilde{\mathbf{y}}_{n,k} = \boldsymbol{\lambda}_{n,k} + \mathbf{n}_{n,k} = \mathbf{g}_\mathrm{fbg}(\mathbf{x}_n(s_{n,k})) + \mathbf{n}_{n,k},
\end{align}
where $\mathbf{g}_\mathrm{fbg}(\mathbf{x}_n(s_{n,k}))$ is a non-linear function relating our state variables to the measured longitudinal strains and $\mathbf{n}_{n,k} \in \mathds{R}^4$ is the sensor noise considering a Gaussian distribution $\mathcal{N}\left( \mathbf{0}, \mathbf{R}_{n,k} \right)$ with covariance matrix $\mathbf{R}_{n,k} \in \mathds{R}^{4\times4}$.
Since the innermost fiber is only affected by elongation deformations, its longitudinal strain is related to our state variables with \cite{Modes2020}
\begin{align}\label{eq:fbg_model_start}
	{\lambda}^{(1)}_{n,k} &= \nu^{(1)}_{n,k} -1,
\end{align}
where $\nu^{(1)}_{n,k}$ is the $x$-component of the translational strain $\mbs{\nu}_n(s_n)$ strain variables, i.e., the elongation strain.
For the remaining three fibers, we can write \cite{Modes2020}
\begin{align}\label{eq:lamba_eq}
	{\lambda}^{(i)}_{n,k} &= \sqrt{\left(\nu^{(1)}_{n,k} - r_n\kappa_n\mathrm{cos}\left(\theta_{n,b} - \theta_{n,i}\right)\right)^2 + \left(r_n\omega^{(1)}_{n,k}\right)^2} - 1,
\end{align}
with $i \in \{2,...,4\}$.
This expression relates the measured longitudinal strain to bending, twisting and elongation deformations.
Here, $\omega^{(1)}_{n,k}$ is the $x$-component of the rotational strain $\mbs{\om}_n(s_n)$, i.e., the twist strain, and $\kappa_n$ and $\theta_{n,b}$ are the overall bending curvature and angle.
Considering the FBG sensor cross-section in Fig.~\ref{fig:fbg_cross_section}, we see that the curvature $\kappa_n$ is equal to the length of the vector of bending strains $\boldsymbol{\omega}_{n,b} = \begin{bmatrix}
	\omega^{(2)}_{n,k} & \omega^{(3)}_{n,k}
\end{bmatrix}^T$.
This allows us to write
\begin{align}
	\omega^{(2)}_{n,k} &= \mathrm{cos}\left(\theta_{n,b} + \frac{\pi}{2}\right)\kappa_n = -\mathrm{sin}\left(\theta_{n,b}\right)\kappa_n, \\
	\omega^{(3)}_{n,k} &= \mathrm{sin}\left(\theta_{n,b} + \frac{\pi}{2}\right)\kappa_n = \mathrm{cos}\left(\theta_{n,b}\right)\kappa_n.
\end{align}
Using this, we can rewrite the expression relying on $\kappa_n$ and $\theta_{n,b}$ in \eqref{eq:lamba_eq} as
\begin{align}
	&r_n\kappa_n\mathrm{cos}\left(\theta_{n,b} - \theta_{n,i}\right), \\ 
	= \hspace{2pt} &r_n\kappa_n\left(\mathrm{cos}\left(\theta_{n,b}\right)\mathrm{cos}\left(\theta_{n,i}\right) + \mathrm{sin}\left(\theta_{n,b}\right)\mathrm{sin}\left(\theta_{n,i}\right)\right), \\
	= \hspace{2pt} &r_n\left(\omega^{(3)}_{n,k}\mathrm{cos}\left(\theta_{n,i}\right) - \omega^{(2)}_{n,k}\mathrm{sin}\left(\theta_{n,i}\right)\right), \label{eq:fbg_model_end}
\end{align}
using trigonometric identities and the fact that $\mathrm{cos}(\theta_{n,b}) = \omega^{(3)}_{n,k}\kappa^{-1}_n$ and $\mathrm{sin}(\theta_{n,b}) = -\omega^{(2)}_{n,k}\kappa^{-1}_n$.

Our non-linear sensor model $\mathbf{g}_\mathrm{fbg}(\mathbf{x}_n(s_{n,k}))$ is now defined using \eqref{eq:fbg_model_start}--\eqref{eq:fbg_model_end} and we can construct a simple error term for our measurements, which is zero when the measured quantities match the ones we would expect from the true state
\begin{equation}\label{eq:strainerror}
	\mathbf{e}_{m,n,k} = \widetilde{\mathbf{y}}_{n,k} - \mathbf{g}_\mathrm{fbg}(\mathbf{x}_n(s_{n,k})).
\end{equation}
The corresponding squared-error cost term results in
\begin{equation}\label{eq:straincost}
	J_{m,n,k} = \frac{1}{2} \mathbf{e}_{m,n,k}^T \mathbf{R}_{n,k}^{-1} \mathbf{e}_{m,n,k}.
\end{equation}

Similar to the prior cost, we sum up the total measurement cost of the system as
\begin{equation}
J_{m} = J_{m,\text{ee}} + \sum_{n=1}^{N} J_{m,n} = J_{m,\text{ee}} + \sum_{n=1}^{N}\sum_{k=0}^{K_n} J_{m,n,k},
\end{equation}
where measurements can include both pose and strain data and terms are dropped for missing measurements.

We note some nuances in the sensor model for parallel optical fibers. 
Firstly, twist direction ambiguity occurs, where both positive and negative twists yield identical sensor readings, leading to potential measurement errors \cite{Modes2020}. 
Secondly, sensitivity to twisting in untwisted configurations is minimal, adversely affecting signal-to-noise ratios and strain reconstruction quality. 
These limitations are well-documented \cite{Modes2020}\cite{Khan2021}\cite{Yi2020}. 
A potential solution is employing helically arranged FBG sensors for the outer fibers, although this complicates manufacturing.

In this paper, we demonstrate that incorporating the derived sensor model into a stochastic state estimation framework, which explicitly accounts for noise, effectively addresses the challenges associated with parallel optical fiber arrangements. 
This approach enhances the interpretation of FBG sensor measurements and their integration with additional state information, including pose measurements, coupling constraints, and prior knowledge, to improve state estimation accuracy.

\subsection{Coupling Cost Terms}	

Lastly, we incorporate coupling constraints into our cost function.
Let us consider that any two poses in our system, including both the continuum robot poses $\mathbf{T}_n(s_{n,k})$ and the end-effector pose $\mathbf{T}_\mathrm{ee}$, can be constrained to each other with a coupling joint $g \in \{1,...,G\}$, where $G$ is the number of total coupling joints in the system.

\begin{figure}[t]
	\centering
	\includegraphics[width=0.5\linewidth]{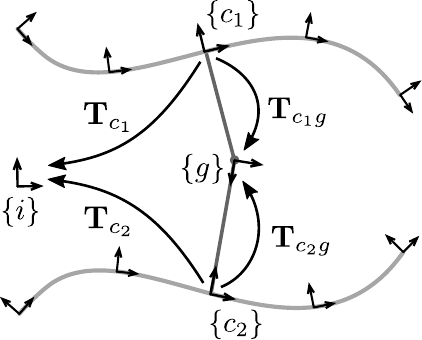}
	\caption{Example of two continuum robots subject to a single coupling constraint. The frames of the coupled poses are denoted $\left\{c_1\right\}$ and $\left\{c_2\right\}$, while the coupling joint frame is denoted with $\{g\}$. Transformation matrices relate each coupled pose to the coupling joint and to the inertial, static frame $\{i\}$.}
	\label{fig:coupling_frames}
\end{figure}

Throughout the following expressions, we will denote the frames of the two coupled poses that a joint $g$ constrains together with $\left\{c_1\right\}$ and $\left\{c_2\right\}$.
Additionally, we denote the frame of the coupling joint itself with $\left\{g\right\}$.
Transformation matrices between the body frames of the coupled poses and the static, inertial frame are denoted as $\mathbf{T}_{c_1}$ and $\mathbf{T}_{c_2}$, while the transformation matrices between these frames and the coupling frames are denoted as $\mathbf{T}_{c_1g}$ and $\mathbf{T}_{c_2g}$.
Both $\mathbf{T}_{c_1}$and $\mathbf{T}_{c_2}$ can directly be obtained from the current state, as they either corresponding to $\mathbf{T}_n(s_{n,k})$ or $\mathbf{T}_\mathrm{ee}$, while $\mathbf{T}_{1g}$ and $\mathbf{T}_{2g}$ are defined by the geometric topology and assembly of the overall system.
An example of two continuum robots that are subject to a single coupling constraint can be seen in Fig.~\ref{fig:coupling_frames}, which additionally visualizes the corresponding frames and transformations.

Assuming the coupling joint $g$ realizes a rigid connection between the two coupled poses in our state, we can define an error term, which expresses how well the coupling constraint is satisfied as
\begin{equation}\label{eq:couplingerr}
\mathbf{e}_{c,g} = \ln \left(\mathbf{T}_{{c_1}g}^{-1}\mathbf{T}_{c_1}\mathbf{T}_{c_2}^{-1}\mathbf{T}_{{c_2}g}\right)^\vee,
\end{equation}
which is a six-dimensional coupling error vector defined with respect to the coupling joint frame $\{g\}$.
If the coupling constraint is exactly satisfied, i.e., the poses of $\mathbf{T}_{c_1}$ and $\mathbf{T}_{c_2}$ exactly `close' this loop of transformations, the error term equals zero.
The squared-error cost term can be constructed analogously to the prior and measurement cost terms as
\begin{equation}
J_{c,g} = \frac{1}{2}\mathbf{e}^T_{c,g}\mathbf{R}_{c,g}^{-1}\mathbf{e}_{c,g}.
\end{equation}
During state estimation, the matrix $\mathbf{R}_{c,g}\in\mathds{R}^{6\times6}$ weighs the coupling constraint of joint $g$ against the prior and measurement cost terms in our overall cost function.
In our approach, we prioritize coupling constraints by assigning them higher weights relative to the prior and measurement terms. 
This weighting can be adjusted to reflect uncertainties in the coupling constraints, like clearance and backlash. 
Additionally, for constraining specific pose aspects, a projection matrix is used to isolate unconstrained degrees of freedom in~\eqref{eq:couplingerr}, accommodating various joint types like spherical joints, which constrain position but not orientation.

Each coupling constraint adds an additional \textit{binary} factor to our factor graph, involving the discrete states subject to coupling (see Fig.~\ref{fig:factor_graph}, right).
The total cost considering all coupling joints and constraints present in the parallel continuum robot can be written as
\begin{equation}
J_{c} = \sum_{g=1}^{G}J_{c,g}.
\end{equation}

\subsection{Overall Objective Function and Batch Formulation}

Putting all of our cost terms together, the overall cost that we seek to minimize is
\begin{equation}\label{eq:totalcost}
J = J_p + J_m + J_c.
\end{equation}
Our MAP optimization problem is defined as
\begin{equation}
\est{\mathbf{x}} = \mbox{arg}\min_{\mathbf{x}} J(\mathbf{x}),
\end{equation}
where $\est{\mathbf{x}}$ is the most likely state of our system taking into account prior knowledge, sensor measurements and coupling constrains.
Throughout the following, we will express our system state $\mathbf{x}$ as a stacked column as
\begin{equation}
\mathbf{x} = \bbm \begin{array}{c} \mathbf{x}_1(s_{1,0}) \\ \mathbf{x}_1(s_{1,1}) \\ \vdots \\ \mathbf{x}_1(s_{1,K_1}) \\ \hline \vdots \\ \hline \mathbf{x}_N(s_{N,0}) \\ \mathbf{x}_N(s_{N,1}) \\ \vdots \\ \mathbf{x}_N(s_{N,K_N}) \\ \hline \mathbf{x}_\text{ee} \end{array} \ebm,
\end{equation}
which includes the states $\mathbf{x}_n(s_{n,k}) = \{\mathbf{T}_n(s_{n,k}), \mbs{\varepsilon}_n(s_{n,k})\}$ of each continuum robot $n$ and the end-effector state $\mathbf{x}_\text{ee} = \mathbf{T}_\text{ee}$.
We note that stacking the states as a column vector is technically abusing notation since $\mathbf{x}_n(s_{n,k}) \in SE(3) \times \mathds{R}^6$ and $\mathbf{x}_\mathrm{ee} \in SE(3)$.
However, it makes the following derivations more readable and the optimization scheme easier to follow.
Additionally, we rewrite our function using stacked quantities as
\begin{equation}\label{eq:totalcost2}
	J = \frac{1}{2} \mathbf{e}_p^T \mathbf{Q}_p^{-1} \mathbf{e}_p + \frac{1}{2} \mathbf{e}_m^T \mathbf{R}_m^{-1} \mathbf{e}_m + \frac{1}{2} \mathbf{e}_c^T \mathbf{R}_c^{-1} \mathbf{e}_c,
\end{equation}
where
\begin{align}
	\mathbf{e}_p &= \bbm \begin{array}{c}
		\mathbf{e}_{p,1,1} \\ \vdots \\ \mathbf{e}_{p,1,K_1} \\ \hline \vdots \\ \hline \mathbf{e}_{p,N,1} \\ \vdots \\ \mathbf{e}_{p,N,K_N} \\ \hline \mathbf{e}_{p,\text{ee}} \end{array} \ebm,
	\mathbf{e}_m = \bbm \begin{array}{c}
		\mathbf{e}_{m,1,0} \\ \vdots \\ \mathbf{e}_{m,1,K_1} \\ \hline \vdots \\ \hline \mathbf{e}_{m,N,0} \\ \vdots \\ \mathbf{e}_{m,N,K_N} \\ \hline \mathbf{e}_{m,\text{ee}} \end{array} \ebm,
	\mathbf{e}_c = \bbm \mathbf{e}_{c,1} \\ \vdots \\ \mathbf{e}_{c,G} \ebm,
\end{align}
and
\begin{align}
	&\mathbf{Q}_p = \mbox{diag}\left( \mathbf{Q}_{1,1}, ..., \mathbf{Q}_{1,K_1}, ..., \mathbf{Q}_{N,1}, ..., \mathbf{Q}_{N,K_N}\right), \\
	&\mathbf{R}_m = \mbox{diag}\left( \mathbf{R}_{1,0}, ..., \mathbf{R}_{1,K_1}, ..., \mathbf{R}_{N,0}, ..., \mathbf{R}_{N,K_N}, \mathbf{R}_\text{ee} \right), \\
	&\mathbf{R}_c = \mbox{diag}\left( \mathbf{R}_{c,1}, ..., \mathbf{R}_{c,G} \right). 
\end{align}
This new expression is equivalent to the summation in~\eqref{eq:totalcost}.

\section{Maximum A Posteriori Optimization}

Since our resulting cost function is non-linear with respect to our system state $\mathbf{x}$, we will solve the MAP optimization problem iteratively using a Gauss-Newton approach.
In the following, we discuss the linearization of our problem, including the derivation of the necessary Jacobian matrices, and how the linearized system can be solved efficiently.

\subsection{Linearization}

During an iteration of our Gauss-Newton approach, we linearize our optimization problem around an operating point, which is usually the state estimate from the previous iteration.
We then solve the linearized problem to update our state estimate for the next iteration.
For linearization, we need to consider perturbations in our state variables.
We  perturb the continuum robot poses $\mathbf{T}_n(s_{n,k})$ and end-effector pose $\mathbf{T}_\mathrm{ee}$ of our system state in an $SE(3)$-sensitive way \cite{Barfoot2024},
\begin{equation}\label{eq:posepert}
\mathbf{T} = \underbrace{\exp\left( \delta\mbs{t}^\wdg \right)}_{\in \; SE(3)} \mathbf{T}_{\rm op},
\end{equation}
where $\delta\mbs{t} \in \mathbb{R}^6$ is the perturbation and $ \mathbf{T}_{\rm op}$ is the pose at our operating point. The perturbations for the continuum robot strain variables can be done in a straightforward vector space manner
\begin{equation}\label{eq:strainpert}
\mbs{\varepsilon}_n(s_{n,k}) = \mbs{\varepsilon}_n(s_{n,k})_{\rm op} + \delta\mbs{\varepsilon}_{n,k},
\end{equation}
where $\delta\mbs{\varepsilon}_{n,k} \in \mathbb{R}^6$ is the perturbation and $\mbs{\varepsilon}(s_{n,k})_{\rm op}$ is the strain at the operating point.
For the continuum robots, we can combine the two parts of the state as
\begin{equation}\label{eq:cr_pert}
\mathbf{x}_n(s_{n,k}) = \mathbf{x}_n(s_{n,k})_{\rm op} + \delta \mathbf{x}_{n,k}, \qquad \delta \mathbf{x}_{n,k} = \bbm \delta\mbs{t}_{n,k} \\ \delta\mbs{\varepsilon}_{n,k} \ebm.
\end{equation}
For the common end-effector, we simply have
\begin{equation}\label{eq:ee_pert}
\mathbf{x}_\text{ee} = \mathbf{x}_\text{ee,op} + \delta \mathbf{x}_\text{ee}, \qquad \delta \mathbf{x}_\text{ee} =  \delta\mbs{t}_\text{ee}.
\end{equation}
Perturbations of the entire system state can now be written as
\begin{equation}\label{eq:total_pert}
\mathbf{x} = \mathbf{x}_{\rm op} + \delta \mathbf{x}, \qquad \delta\mathbf{x} = \bbm \begin{array}{c} \delta\mathbf{x}_1(s_{1,0}) \\ \vdots \\ \delta\mathbf{x}_1(s_{1,K_1}) \\ \hline \vdots \\ \hline \delta\mathbf{x}_N(s_{N,0}) \\ \vdots \\ \delta\mathbf{x}_N(s_{N,K_N}) \\ \hline \delta\mathbf{x}_\text{ee} \end{array} \ebm.
\end{equation}
We note that in \eqref{eq:cr_pert}, \eqref{eq:ee_pert} and \eqref{eq:total_pert} we again slightly abuse notation as a shorthand. During implementation, perturbations to pose and strain variables are applied according to \eqref{eq:posepert} and \eqref{eq:strainpert}, respectively.
Considering these perturbation schemes, we can now linearize the prior, measurement and coupling errors as
\begin{align}\label{eq:linerrors}
\mathbf{e}_p &= \mathbf{e}_{p,{\rm op}} + \mathbf{E}_p \, \delta\mathbf{x}, \\ \mathbf{e}_m &= \mathbf{e}_{m,{\rm op}} + \mathbf{E}_m \, \delta\mathbf{x}, \\ \mathbf{e}_c &= \mathbf{e}_{c,{\rm op}} + \mathbf{E}_c \, \delta\mathbf{x}.
\end{align}
Here, $\mathbf{e}_{p,{\rm op}}$, $\mathbf{e}_{m,{\rm op}}$ and $\mathbf{e}_{c,{\rm op}}$ are the errors evaluated at our operation point, i.e., using the state estimate of the previous iteration, and $\mathbf{E}_p$, $\mathbf{E}_m$ and $\mathbf{E}_c$ are the error Jacobians.
Substituting the linearized error terms into our overall cost function \eqref{eq:totalcost2} yields
\begin{align}
J \approx \quad &\frac{1}{2} \left( \mathbf{e}_{p,{\rm op}} + \mathbf{E}_p \, \delta\mathbf{x} \right)^T \mathbf{Q}_p^{-1} \left( \mathbf{e}_{p,{\rm op}} + \mathbf{E}_p \, \delta\mathbf{x} \right) \notag \\
+ &\frac{1}{2} \left( \mathbf{e}_{m,{\rm op}} + \mathbf{E}_m \, \delta\mathbf{x} \right)^T \mathbf{R}_m^{-1} \left( \mathbf{e}_{m,{\rm op}} + \mathbf{E}_m \, \delta\mathbf{x} \right) \\
+ &\frac{1}{2} \left( \mathbf{e}_{c,{\rm op}} + \mathbf{E}_c \, \delta\mathbf{x} \right)^T \mathbf{R}_c^{-1} \left( \mathbf{e}_{c,{\rm op}} + \mathbf{E}_c \, \delta\mathbf{x} \right) \notag,
\end{align}
which is quadratic w.r.t. our perturbation variable, $\delta\mathbf{x}$.
We can rewrite this expression into a linear system of equations
\begin{align}\label{eq:linsys}
\biggl( \underbrace{\mathbf{E}_p^T \mathbf{Q}_p^{-1} \mathbf{E}_p + \mathbf{E}_m^T \mathbf{R}_m^{-1} \mathbf{E}_m}_{\text{block-tridiagonal}}  + \underbrace{\mathbf{E}_c^T \mathbf{R}_c^{-1} \mathbf{E}_c}_{\text{off-diagonal~entries}} \biggr) \, \delta\mathbf{x}^\star = \hfill \notag\\
- \left( \mathbf{E}_p^T \mathbf{Q}_p^{-1}  \mathbf{e}_{p,{\rm op}} + \mathbf{E}_m^T \mathbf{R}_m^{-1} \mathbf{e}_{m,{\rm op}} + \mathbf{E}_c^T \mathbf{R}_c^{-1} \mathbf{e}_{c,{\rm op}} \right),
\end{align}
where $\delta\mathbf{x}^\star$ is the state perturbation that minimizes our linearized problem.
The matrix of our linearized system features a particular sparsity pattern.
This system matrix contains block-tridiagonal entries resulting from our prior and measurement cost terms, while the coupling cost terms lead to additional off-diagonal entries.

\subsection{Jacobian Matrices}

In the following, we derive expressions for the error Jacobians $\mathbf{E}_p$, $\mathbf{E}_m$, and $\mathbf{E}_c$, which are required to solve~\eqref{eq:linsys}. 

\subsubsection{Prior}
As shown in our prior work \cite{Lilge2022}, the prior error of each continuum robot can be linearized as,
\begin{align}
	\mathbf{e}_{p,n,k} &= \mathbf{e}_{p,n,k,{\rm op}} + \mathbf{E}_{p,n,k}\bbm \delta\mbs{t}_{n,k-1} \\ \delta\mbs{\varepsilon}_{n,k-1} \\ \delta\mbs{t}_{n,k} \\ \delta\mbs{\varepsilon}_{n,k}\ebm, \notag \\
\end{align}
with
\begin{align}
	\mathbf{E}_{p,n,k} &= \bbm -\mbs{\mathcal{J}}_{\rm op}^{-1} \mbs{\mathcal{T}}_{\rm op} & - \Delta s_{n,k} \mathbf{1} & \mbs{\mathcal{J}}_{\rm op}^{-1} & \mathbf{0} \\ -\mbs{\bar{\mathcal{J}}}_{\rm op}^{-1} \mbs{\mathcal{T}}_{\rm op} & - \mathbf{1} & \mbs{\bar{\mathcal{J}}}_{\rm op}^{-1} & \mbs{\mathcal{J}}_{\rm op}^{-1} \ebm ,
\end{align}
where
\begin{align}
	\Delta s_{n,k} &= (s_{n,k} - s_{n,k-1}), \\
	\mbs{\mathcal{J}}_{\rm op}^{-1} &= \mbs{\mathcal{J}}\left( \ln \left(   \mathbf{T}_n(s_{n,k})_{\rm op} \mathbf{T}_n(s_{n,k-1})_{\rm op}^{-1} \right)^\vee  \right)^{-1}, \\
	\mbs{\mathcal{T}}_{\rm op} &= \mbox{Ad}\left( \mathbf{T}_n(s_{n,k})_{\rm op} \mathbf{T}_n(s_{n,k-1})_{\rm op}^{-1} \right), \\
	\mbs{\bar{\mathcal{J}}}_{\rm op}^{-1} &= \frac{1}{2} \mbs{\varepsilon}_{n,k,{\rm op}}^\Wdg \mbs{\mathcal{J}}^{-1}. 
\end{align}
Depending on the indices $n$ and $k$ of the variables involved in each individual expression, the subblocks of $\mathbf{E}_{p,n,k}$ must be placed into the corresponding entries of the overall Jacobian, $\mathbf{E}_p$.

\subsubsection{Measurements}
For measurements of continuum robot poses we have \cite{Lilge2022}
\begin{align}
	\mathbf{e}_{m,n,k} = \mathbf{e}_{m,n,k,{\rm op}} +\hspace{4cm}& \notag \\ \underbrace{\bbm \mbs{\mathcal{J}}\left(  \ln \left(   \mathbf{T}_n(s_{n,k})_{\rm op} \widetilde{\mathbf{T}}_{n,k}^{-1} \right)^\vee \right)^{-1} & \mathbf{0} \ebm}_{\mathbf{E}_{m,n,k}} & \bbm \delta\mbs{t}_{n,k} \\ \delta \mbs{\varepsilon}_{n,k} \ebm.
\end{align}
The error for an end-effector pose measurement can be linearized analogously as
\begin{equation}
	\mathbf{e}_{m,\text{ee}} = \mathbf{e}_{m,{\rm ee,op}} + \underbrace{\bbm \mbs{\mathcal{J}}\left(  \ln \left(   \mathbf{T}_{\rm ee,op} \widetilde{\mathbf{T}}_\text{ee}^{-1} \right)^\vee \right)^{-1} \ebm}_{\mathbf{E}_{m,\text{ee}}}  \delta\mbs{t}_{\text{ee}}.
\end{equation}

Linearizing the error terms for our strain measurements is not straight-forward, as the sensor model is non-linear $\mathbf{g}_\mathrm{fbg}(\mathbf{x}_n(s_{n,k}))$.
Considering the error term \eqref{eq:strainerror}, we have
\begin{equation}
	\mathbf{e}_{m,n,k} = \mathbf{e}_{m,n,k,{\rm op}}~\underbrace{- \frac{\partial\mathbf{g}_\mathrm{fbg}(\mathbf{x}_n(s_{n,k}))}{\partial\mathbf{x}_n(s_{n,k})}}_{\mathbf{E}_{m,n,k}} \bbm \delta\mbs{t}_{n,k} \\ \delta \mbs{\varepsilon}_{n,k} \ebm.
\end{equation}
Considering the pose and strain parts separately, deriving our sensor model with respect to our state variables results in
\begin{align}
	\frac{\partial\mathbf{g}_\mathrm{fbg}}{\partial\mbs{t}_{n,k}} &= \mathbf{0}, \\ 
	\frac{\partial\mathbf{g}_\mathrm{fbg}}{\partial\mbs{\varepsilon}_{n,k}} &= \renewcommand\arraycolsep{3pt} \renewcommand\arraystretch{2.5} \bbm 1 & 0 & 0 & 0 & 0 & 0 \\  \frac{G^{(2)}_1}{{\lambda}^{(2)}_{n,k} + 1} & 0 & 0 & \frac{\omega^{(1)}_{n,k}r^2_n}{{\lambda}^{(2)}_{n,k} + 1} & \frac{r_n\mathrm{s}_{n,2}G^{(2)}_1}{{\lambda}^{(2)}_{n,k} + 1} & \frac{-r_n\mathrm{c}_{n,2}G^{(2)}_1}{{\lambda}^{(2)}_{n,k} + 1} \\ \frac{G^{(3)}_1}{{\lambda}^{(3)}_{n,k} + 1} & 0 & 0 & \frac{\omega^{(1)}_{n,k}r^2_n}{{\lambda}^{(3)}_{n,k} + 1} & \frac{r_n\mathrm{s}_{n,3}G^{(3)}_1}{{\lambda}^{(3)}_{n,k} + 1} & \frac{-r_n\mathrm{c}_{n,3}G^{(3)}_1}{{\lambda}^{(3)}_{n,k} + 1} \\ \frac{G^{(4)}_1}{{\lambda}^{(4)}_{n,k} + 1} & 0 & 0 & \frac{\omega^{(1)}_{n,k}r^2_n}{{\lambda}^{(4)}_{n,k} + 1} & \frac{r_n\mathrm{s}_{n,4}G^{(4)}_1}{{\lambda}^{(4)}_{n,k} + 1} & \frac{-r_n\mathrm{c}_{n,4}G^{(4)}_1}{{\lambda}^{(4)}_{n,k} + 1}\ebm,
\end{align}
with
\begin{align}
	\mathrm{c}_{n,i} &= \mathrm{cos}(\theta_{n,i}), \\
	\mathrm{s}_{n,i} &= \mathrm{sin}(\theta_{n,i}), \\ 
	G^{(i)}_1 &= \nu^{(1)}_{n,k} - r_n\left(\omega^{(3)}_{n,k}\mathrm{cos}\left(\theta_{n,i}\right) - \omega^{(2)}_{n,k}\mathrm{sin}\left(\theta_{n,i}\right)\right).
\end{align}
With this, the error Jacobian for our strain measurements is
\begin{align}
	\mathbf{E}_{m,n,k} = \bbm \mathbf{0} & -\frac{\partial\mathbf{g}_\mathrm{fbg}}{\partial\mbs{\varepsilon}_{n,k}} \ebm.
\end{align}
The Jacobian of our sensor model becomes rank-deficient when $\omega^{(1)}_{n,k} = 0$, indicating no twisting deformations in the sensor error. 
Near this singularity, FBG sensor readings barely respond to twists, reducing sensitivity and potentially compromising accurate twist determination in continuum robots. 
To mitigate this, fusing the FBG sensor data with other state estimation components like the prior, pose measurements, and coupling information is effective.

As with the prior terms, the blocks of $\mathbf{E}_{m,n,k}$ and $\mathbf{E}_{m,\text{ee}}$ must be placed into the appropriate blocks of the overall Jacobian, $\mathbf{E}_m$, according to the indices $n$ and $k$ of the variables involved.

\begin{figure*}[!t]
	\begin{tabular}{ccc}
		\parbox{4.5cm}{\centering \textbf{Robot Topology \\ \hspace{1cm}}} & \parbox{6cm}{\centering \textbf{Linear System \\ (Left-Hand Coefficient Matrix)}} & \parbox{6cm}{\centering \textbf{Lower Triangular Matrix of \\ Cholesky Decomposition}} \\
		&&\\
		\parbox{4.5cm}{} & \parbox{6cm}{\centering \small Robot 1 \hspace{1.4cm} Robot 2} & \parbox{6cm}{\centering \small Robot 1 \hspace{1.4cm} Robot 2} \\
		\parbox{4.5cm}{\hfill\includegraphics[width=0.25\textwidth]{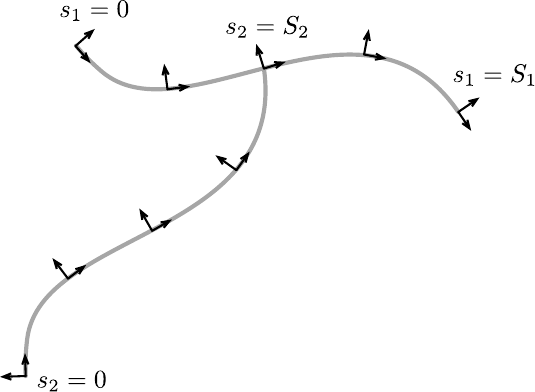}} & \parbox{6cm}{\begin{equation}
				\tiny
				\setlength\arraycolsep{2pt}
				\begin{bmatrix}\renewcommand*{\arraystretch}{1.2}
					\begin{array}{cc;{2pt/6pt}cc;{2pt/6pt}cc;{2pt/6pt}cc;{2pt/6pt}cc|cc;{2pt/6pt}cc;{2pt/6pt}cc;{2pt/6pt}cc;{2pt/6pt}ccc}
						\bullet & \bullet & \bullet & \bullet &  &  &  &  &  &  &  &  &  &  &  &  &  &  &  &  \\
						\bullet & \bullet & \bullet & \bullet &  &  &  &  &  &  &  &  &  &  &  &  &  &  &  &  \\ \hdashline[2pt/6pt]
						\bullet & \bullet & \bullet & \bullet & \bullet & \bullet &  &  &  &  &  &  &  &  &  &  &  &  &  &  \\
						\bullet & \bullet & \bullet & \bullet & \bullet & \bullet &  &  &  &  &  &  &  &  &  &  &  &  &  &  \\ \hdashline[2pt/6pt]
						&  & \bullet & \bullet & \bullet & \bullet & \bullet & \bullet &  &  &  &  &  &  &  &  &  &  & \bullet &  \\
						&  & \bullet & \bullet & \bullet & \bullet & \bullet & \bullet &  &  &  &  &  &  &  &  &  &  &  &  \\ \hdashline[2pt/6pt]
						&  &  &  & \bullet & \bullet & \bullet & \bullet & \bullet & \bullet &  &  &  &  &  &  &  &  &  &  \\
						&  &  &  & \bullet & \bullet & \bullet & \bullet & \bullet & \bullet &  &  &  &  &  &  &  &  &  &  \\ \hdashline[2pt/6pt]
						&  &  &  &  &  & \bullet & \bullet & \bullet & \bullet &  &  &  &  &  &  &  &  &  &  \\
						&  &  &  &  &  & \bullet & \bullet & \bullet & \bullet &  &  &  &  &  &  &  &  &  &  \\ \hline
						&  &  &  &  &  &  &  &  &  & \bullet & \bullet & \bullet & \bullet &  &  &  &  &  &  \\
						&  &  &  &  &  &  &  &  &  & \bullet & \bullet & \bullet & \bullet &  &  &  &  &  &  \\ \hdashline[2pt/6pt]
						&  &  &  &  &  &  &  &  &  & \bullet & \bullet & \bullet & \bullet & \bullet & \bullet &  &  &  &  \\
						&  &  &  &  &  &  &  &  &  & \bullet & \bullet & \bullet & \bullet & \bullet & \bullet &  &  &  &  \\ \hdashline[2pt/6pt]
						&  &  &  &  &  &  &  &  &  &  &  & \bullet & \bullet & \bullet & \bullet & \bullet & \bullet &  &  \\
						&  &  &  &  &  &  &  &  &  &  &  & \bullet & \bullet & \bullet & \bullet & \bullet & \bullet &  &  \\ \hdashline[2pt/6pt]
						&  &  &  &  &  &  &  &  &  &  &  &  &  & \bullet & \bullet & \bullet & \bullet & \bullet & \bullet \\
						&  &  &  &  &  &  &  &  &  &  &  &  &  & \bullet & \bullet & \bullet & \bullet & \bullet & \bullet \\  \hdashline[2pt/6pt]
						&  &  &  & \bullet &  &  &  &  &  &  &  &  &  &  &  & \bullet & \bullet & \bullet & \bullet \\
						&  &  &  &  &  &  &  &  &  &  &  &  &  &  &  & \bullet & \bullet & \bullet & \bullet  
					\end{array}
				\end{bmatrix} \notag
		\end{equation}} & \parbox{6cm}{\begin{equation}
				\tiny
				\setlength\arraycolsep{2pt}
				\begin{bmatrix}\renewcommand*{\arraystretch}{1.2}
					\begin{array}{cc;{2pt/6pt}cc;{2pt/6pt}cc;{2pt/6pt}cc;{2pt/6pt}cc|cc;{2pt/6pt}cc;{2pt/6pt}cc;{2pt/6pt}cc;{2pt/6pt}ccc}
						\bullet &  &  &  &  &  &  &  &  &  &  &  &  &  &  &  &  &  &  &  \\
						\bullet & \bullet &  &  &  &  &  &  &  &  &  &  &  &  &  &  &  &  &  &  \\ \hdashline[2pt/6pt]
						\bullet & \bullet & \bullet &  &  &  &  &  &  &  &  &  &  &  &  &  &  &  &  &  \\
						\bullet & \bullet & \bullet & \bullet &  &  &  &  &  &  &  &  &  &  &  &  &  &  &  &  \\ \hdashline[2pt/6pt]
						&  & \bullet & \bullet & \bullet &  &  &  &  &  &  &  &  &  &  &  &  &  &  &  \\
						&  & \bullet & \bullet & \bullet & \bullet &  &  &  &  &  &  &  &  &  &  &  &  &  &  \\ \hdashline[2pt/6pt]
						&  &  &  & \bullet & \bullet & \bullet &  &  &  &  &  &  &  &  &  &  &  &  &  \\
						&  &  &  & \bullet & \bullet & \bullet & \bullet &  &  &  &  &  &  &  &  &  &  &  &  \\ \hdashline[2pt/6pt]
						&  &  &  &  &  & \bullet & \bullet & \bullet &  &  &  &  &  &  &  &  &  &  &  \\
						&  &  &  &  &  & \bullet & \bullet & \bullet & \bullet &  &  &  &  &  &  &  &  &  &  \\ \hline
						&  &  &  &  &  &  &  &  &  & \bullet &  &  &  &  &  &  &  &  &  \\
						&  &  &  &  &  &  &  &  &  & \bullet & \bullet &  &  &  &  &  &  &  &  \\ \hdashline[2pt/6pt]
						&  &  &  &  &  &  &  &  &  & \bullet & \bullet & \bullet &  &  &  &  &  &  &  \\
						&  &  &  &  &  &  &  &  &  & \bullet & \bullet & \bullet & \bullet &  &  &  &  &  &  \\ \hdashline[2pt/6pt]
						&  &  &  &  &  &  &  &  &  &  &  & \bullet & \bullet & \bullet &  &  &  &  &  \\
						&  &  &  &  &  &  &  &  &  &  &  & \bullet & \bullet & \bullet & \bullet &  &  &  &  \\ \hdashline[2pt/6pt]
						&  &  &  &  &  &  &  &  &  &  &  &  &  & \bullet & \bullet & \bullet &  &  &  \\
						&  &  &  &  &  &  &  &  &  &  &  &  &  & \bullet & \bullet & \bullet & \bullet &  &  \\  \hdashline[2pt/6pt]
						&  &  &  & \bullet & \boldsymbol{\circ} & \boldsymbol{\circ} & \boldsymbol{\circ} & \boldsymbol{\circ} & \boldsymbol{\circ} &  &  &  &  &  &  & \bullet & \bullet & \bullet &  \\
						&  &  &  &  &  &  &  &  &  &  &  &  &  &  &  & \bullet & \bullet & \bullet & \bullet 
					\end{array}
				\end{bmatrix} \notag
		\end{equation}} \\
		& \small \hspace*{-3.07cm} \parbox{3cm}{ \centering \vspace*{-1cm}  \large $\uparrow$} &  \hspace*{-1.45cm} \parbox{3cm}{ \centering \vspace*{-1cm}  \large $\underbrace{\hspace{1.25cm}}$} \\
		 & \small \hspace*{-3.07cm} \parbox{3cm}{ \centering \vspace*{-1cm}  {\small Coupling}} &  \hspace*{-1.45cm} \parbox{3cm}{ \centering \vspace*{-1cm} {\small Coupling Fill-In}} \\
		\parbox{4.5cm}{\hfill\includegraphics[width=0.25\textwidth]{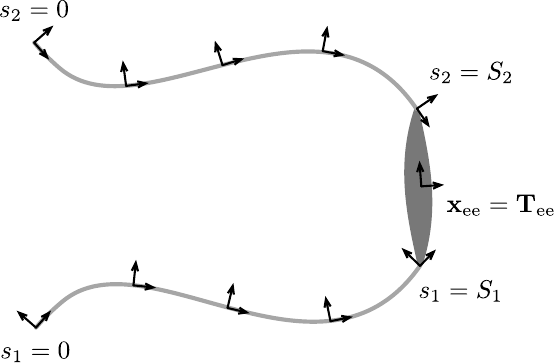}} & \parbox{6cm}{\begin{equation}
				\tiny
				\setlength\arraycolsep{2pt}
				\begin{bmatrix}\renewcommand*{\arraystretch}{1.2}
					\begin{array}{cc;{2pt/6pt}cc;{2pt/6pt}cc;{2pt/6pt}cc;{2pt/6pt}cc|cc;{2pt/6pt}cc;{2pt/6pt}cc;{2pt/6pt}cc;{2pt/6pt}cc|c}
						\bullet & \bullet & \bullet & \bullet &  &  &  &  &  &  &  &  &  &  &  &  &  &  &  & & \\
						\bullet & \bullet & \bullet & \bullet &  &  &  &  &  &  &  &  &  &  &  &  &  &  &  & & \\ \hdashline[2pt/6pt]
						\bullet & \bullet & \bullet & \bullet & \bullet & \bullet &  &  &  &  &  &  &  &  &  &  &  &  &  & & \\
						\bullet & \bullet & \bullet & \bullet & \bullet & \bullet &  &  &  &  &  &  &  &  &  &  &  &  &  & & \\ \hdashline[2pt/6pt]
						&  & \bullet & \bullet & \bullet & \bullet & \bullet & \bullet &  &  &  &  &  &  &  &  &  &  &  & & \\
						&  & \bullet & \bullet & \bullet & \bullet & \bullet & \bullet &  &  &  &  &  &  &  &  &  &  &  & & \\ \hdashline[2pt/6pt]
						&  &  &  & \bullet & \bullet & \bullet & \bullet & \bullet & \bullet &  &  &  &  &  &  &  &  &  & & \\
						&  &  &  & \bullet & \bullet & \bullet & \bullet & \bullet & \bullet &  &  &  &  &  &  &  &  &  & & \\ \hdashline[2pt/6pt]
						&  &  &  &  &  & \bullet & \bullet & \bullet & \bullet &  &  &  &  &  &  &  &  &  & & \bullet \\
						&  &  &  &  &  & \bullet & \bullet & \bullet & \bullet &  &  &  &  &  &  &  &  &  & & \\ \hline
						&  &  &  &  &  &  &  &  &  & \bullet & \bullet & \bullet & \bullet &  &  &  &  &  & & \\
						&  &  &  &  &  &  &  &  &  & \bullet & \bullet & \bullet & \bullet &  &  &  &  &  & & \\ \hdashline[2pt/6pt]
						&  &  &  &  &  &  &  &  &  & \bullet & \bullet & \bullet & \bullet & \bullet & \bullet &  &  &  & & \\
						&  &  &  &  &  &  &  &  &  & \bullet & \bullet & \bullet & \bullet & \bullet & \bullet &  &  &  & & \\ \hdashline[2pt/6pt]
						&  &  &  &  &  &  &  &  &  &  &  & \bullet & \bullet & \bullet & \bullet & \bullet & \bullet &  & & \\
						&  &  &  &  &  &  &  &  &  &  &  & \bullet & \bullet & \bullet & \bullet & \bullet & \bullet &  & & \\ \hdashline[2pt/6pt]
						&  &  &  &  &  &  &  &  &  &  &  &  &  & \bullet & \bullet & \bullet & \bullet & \bullet & \bullet & \\
						&  &  &  &  &  &  &  &  &  &  &  &  &  & \bullet & \bullet & \bullet & \bullet & \bullet & \bullet & \\  \hdashline[2pt/6pt]
						&  &  &  &  &  &  &  &  &  &  &  &  &  &  &  & \bullet & \bullet & \bullet & \bullet & \bullet \\
						&  &  &  &  &  &  &  &  &  &  &  &  &  &  &  & \bullet & \bullet & \bullet & \bullet & \\ \hline
						&  &  &  &  &  &  &  & \bullet &  &  &  &  &  &  &  &  &  & \bullet &  & \bullet \\
					\end{array}
				\end{bmatrix} \notag
		\end{equation}} & \parbox{6cm}{\begin{equation}
				\tiny
				\setlength\arraycolsep{2pt}
				\begin{bmatrix}\renewcommand*{\arraystretch}{1.2}
					\begin{array}{cc;{2pt/6pt}cc;{2pt/6pt}cc;{2pt/6pt}cc;{2pt/6pt}cc|cc;{2pt/6pt}cc;{2pt/6pt}cc;{2pt/6pt}cc;{2pt/6pt}cc|c}
						\bullet &  &  &  &  &  &  &  &  &  &  &  &  &  &  &  &  &  &  & & \\
						\bullet & \bullet &  &  &  &  &  &  &  &  &  &  &  &  &  &  &  &  &  & & \\ \hdashline[2pt/6pt]
						\bullet & \bullet & \bullet &  & & &  &  &  &  &  &  &  &  &  &  &  &  &  & & \\
						\bullet & \bullet & \bullet & \bullet & &  &  &  &  &  &  &  &  &  &  &  &  &  &  & & \\
						\hdashline[2pt/6pt]&  & \bullet & \bullet & \bullet &  &  & &  &  &  &  &  &  &  &  &  &  &  & & \\
						&  & \bullet & \bullet & \bullet & \bullet &  &  &  &  &  &  &  &  &  &  &  &  &  & & \\ \hdashline[2pt/6pt]
						&  &  &  & \bullet & \bullet & \bullet &  &  &  &  &  &  &  &  &  &  &  &  & & \\
						&  &  &  & \bullet & \bullet & \bullet & \bullet &  &  &  &  &  &  &  &  &  &  &  & & \\ \hdashline[2pt/6pt]
						&  &  &  &  &  & \bullet & \bullet & \bullet & &  &  &  &  &  &  &  &  &  & & \\
						&  &  &  &  &  & \bullet & \bullet & \bullet & \bullet &  &  &  &  &  &  &  &  &  & & \\ \hline
						&  &  &  &  &  &  &  &  &  & \bullet & &  &  &  &  &  &  &  & & \\
						&  &  &  &  &  &  &  &  &  & \bullet & \bullet &  &  &  &  &  &  &  & & \\ \hdashline[2pt/6pt]
						&  &  &  &  &  &  &  &  &  & \bullet & \bullet & \bullet &&  &  &  &  &  & & \\
						&  &  &  &  &  &  &  &  &  & \bullet & \bullet & \bullet & \bullet &  &  &  &  &  & & \\ \hdashline[2pt/6pt]
						&  &  &  &  &  &  &  &  &  &  &  & \bullet & \bullet & \bullet &  &  &  &  & & \\
						&  &  &  &  &  &  &  &  &  &  &  & \bullet & \bullet & \bullet & \bullet &  & &  & & \\ \hdashline[2pt/6pt]
						&  &  &  &  &  &  &  &  &  &  &  &  &  & \bullet & \bullet & \bullet & &  &  & \\
						&  &  &  &  &  &  &  &  &  &  &  &  &  & \bullet & \bullet & \bullet & \bullet &  &  & \\  \hdashline[2pt/6pt]
						&  &  &  &  &  &  &  &  &  &  &  &  &  &  &  & \bullet & \bullet & \bullet & &  \\
						&  &  &  &  &  &  &  &  &  &  &  &  &  &  &  & \bullet & \bullet & \bullet & \bullet & \\ \hline
						&  &  &  &  &  &  &  & \bullet & \boldsymbol{\circ} &  &  &  &  &  &  &  &  & \bullet & \boldsymbol{\circ} & \bullet \\
					\end{array}
				\end{bmatrix} \notag
		\end{equation}} \\
	\end{tabular}
	\caption{Sparsity patterns of the linear system and the lower triangular matrix of the Cholesky decomposition for two system topologies.
	Each $\bullet$-symbol denotes a $6\times6$ matrix block. As indicated by the dashed lines, four of these blocks make up one discrete node/state along the length of a continuum robot, including pose and strain. Solid lines indicate the separation of states between the individual parts of the system, i.e., continuum robots and end-effector. Sparsity fill-ins of the Cholesky decomposition are indicated by $\boldsymbol{\circ}$-symbols.}
	\label{fig:sparsity}
\end{figure*}

\subsubsection{Coupling}

Lastly, we linearize the error for the coupling constraint enforced by joint $g$ between two individual poses $\mathbf{T}_{c_1}$ and $\mathbf{T}_{c_2}$ as
\begin{align}
	\mathbf{e}_{c,g} &= \mathbf{e}_{c,g,{\rm op}} + \underbrace{\bbm \mathbf{E}_{c,g,1} & \mathbf{E}_{c,g,2} \ebm}_{\mathbf{E}_{c,g}}  \bbm \delta\mbs{t}_{c_1} \\ \delta \mbs{t}_{c_2} \ebm,  
\end{align}
with
\begin{align}
	\mathbf{E}_{c,g,1} &= \mbs{\mathcal{J}}_{g,{c_1},{c_2}}^{-1} \mbox{Ad}\left( \mathbf{T}_{{c_1}g}^{-1} \right) , \\
	\mathbf{E}_{c,g,2} &=  -\mbs{\mathcal{J}}_{g,{c_1},{c_2}}^{-1} \mbox{Ad}\left( \mathbf{T}_{{c_1}g}^{-1}\mathbf{T}_{c_1}\mathbf{T}_{c_2}^{-1} \right),
\end{align}
and
\begin{align}
	\mbs{\mathcal{J}}_{g,1,2} = \mbs{\mathcal{J}}\left(  \ln \left(\mathbf{T}_{1g}^{-1}\mathbf{T}_1\mathbf{T}_2^{-1}\mathbf{T}_{2g}\right)^\vee \right).
\end{align}
Here, $\mathbf{E}_{c,g}$ consists of two parts, $\mathbf{E}_{c,g,1}$ and $\mathbf{E}_{c,g,2}$, each relating changes to the coupling constraint error of joint $g$ to changes in poses $\mathbf{T}_{c_1}$ and $\mathbf{T}_{c_1}$, expressed by $\delta \mbs{t}_{c_1}$ and $\delta \mbs{t}_{c_1}$.
For each coupling joint $g$, the two blocks $\mathbf{E}_{c,g,1}$ and $\mathbf{E}_{c,g,2}$ must be placed into the appropriate blocks of the overall Jacobian $\mathbf{E}_c$, according to the poses involved.

\subsection{Solving the Linear System}

Our linear system predominantly exhibits a block-tridiagonal pattern due to the binary and unary factors of the prior and measurement terms. 
Coupling terms, resulting from joints that link certain poses, introduce additional off-diagonal entries. 
However, given the typically low number of these couplings in state-of-the-art continuum robots (usually one per robot), the system retains an overall sparse structure. 
This sparsity, predetermined by the system's topology, allows efficient solution in $O\left(N\sum_{n=1}^{N}K_n\right)$ using methods like sparse Cholesky decomposition \cite{Meurant1992}, with minimal sparsity fill-in that can be predicted as outlined in \cite{Erisman1975}. 

Figure~\ref{fig:sparsity} (top) shows the sparsity patterns for the linear system and for the lower triangular matrix of the Cholesky decomposition for two coupled continuum robots.
Here, the tip (distal end) of one continuum robot is coupled to the body of the second.
This coupling constraint leads to off-diagonal entries in addition to the block-tridiagonal structure arising from our prior and measurement terms.
These off-diagonal entries further lead to a fill-in in the Cholesky decomposition.
However, the fill-in is minimal and can be determined using \cite{Erisman1975}, leading to the depicted sparsity pattern of the lower triangular matrix of the Cholesky decomposition.

Figure~\ref{fig:sparsity} (bottom) illustrates the sparsity patterns for two continuum robots coupled to a common end-effector.
Similar to the above, we observe both a block-tridiagonal structure of the linear system from the prior and measurement terms with additional off-diagonal entries form the coupling constraints.
This again leads to fill-ins in the sparsity pattern of the depicted lower triangular matrix of the Cholesky decomposition.

The sparsity fill-in during Cholesky decomposition is significantly influenced by the ordering of variables in the state vector, particularly the arrangement of individual continuum robots' states. 
While finding the optimal order for minimal fill-in is trivial for some problems, like a system with two robots coupled to a common end-effector, it can be improved in more complex configurations, such as a robot with its tip coupled to another's body, by reordering the states. 
Practically, it's essential to arrange the states in coupled systems to minimize fill-in. 
Identifying the optimal order is an NP-hard problem \cite{Yannakakis1981}, good heuristics are a practical solution (see \cite[Chapter 3]{Meurant1999} for examples).

Lastly, it should be emphasized that the resulting system matrix needs to exhibit full rank in order to solve the state estimation problem, i.e., the system needs to be fully defined and constrained.
This can be ensured by adding sufficient coupling constraints and measurements, or by enforcing boundary conditions, which we will discuss below.

\subsection{Updating the State}

After solving for the optimal perturbation $\delta\mathbf{x}^*$ of our state for the current iteration, the operating points of our state variables can be updated using
\begin{align}\mathbf{T}_n(s_{n,k})_{\rm op} & \leftarrow  \exp\left( \alpha \, \delta\mbs{t}_{n,k}^{\star^\wdg} \right) \mathbf{T}_n(s_{n,k})_{\rm op}, \\
	\mbs{\varepsilon}_n(s_{n,k})_{\rm op} & \leftarrow  \mbs{\varepsilon}_n(s_{n,k})_{\rm op} + \alpha \, \delta\mbs{\varepsilon}_{n,k}^\star, \\
	\mathbf{T}_{\rm ee,op} & \leftarrow  \exp\left( \alpha \, \delta\mbs{t}_\text{ee}^{\star^\wdg} \right) \mathbf{T}_{\rm ee,op}.
\end{align}
We apply a step of size $\alpha$ along the computed perturbation $\delta\mathbf{x}^*$. 
Practically, a line-search method can be used to determine an optimal $\alpha$ to minimize cost between iterations. 
The presented optimization scheme iterates until $\delta\mathbf{x}^*$ is sufficiently small, indicating convergence.

In the optimization process, a projection matrix can be used to selectively mask certain state system entries, keeping them constant at initial values. 
This technique enforces boundary conditions, such as fixing the proximal/distal poses or strains of continuum robots, or enforcing Kirchhoff rod-like behaviour by locking translational strains to $\mbs{\nu}_n(s_{n,k})=\begin{bmatrix} 1 & 0 & 0 \end{bmatrix}^T$.

\subsection{Obtaining the Posterior Distribution}

Once our optimization converges, the operating point of our last iteration will be the most likely state of our state estimate.
We can obtain the uncertainty of this estimate by taking the left-hand side of the linear system of equations~\eqref{eq:linsys} at the last iteration and invert this for the covariance, $\hat{\mathbf{P}}$:
\begin{equation}\label{eq:fullcov}
\hat{\mathbf{P}} = \left(  \mathbf{E}_p^T \mathbf{Q}^{-1} \mathbf{E}_p + \mathbf{E}_m^T \mathbf{R}_m^{-1} \mathbf{E}_m + \mathbf{E}_c^T \mathbf{R}_c^{-1} \mathbf{E}_c \right)^{-1}.
\end{equation}
In practice, we are only interested in certain subblocks of $\hat{\mathbf{P}}$ and we are referring to our prior work for more detail on how to obtain those \cite{Lilge2022}.
Additionally, we can use the Gaussian process interpolation equations to query the mean and covariance of the continuum robot states at any continuous arclength $s_n$.
This can be done in $O(1)$ time and we again refer to \cite{Lilge2022} for additional details.

\begin{figure*}[!t]
	\centering
	\scriptsize
	\def\svgwidth{1\linewidth}
\begingroup%
  \makeatletter%
  \providecommand\color[2][]{%
    \errmessage{(Inkscape) Color is used for the text in Inkscape, but the package 'color.sty' is not loaded}%
    \renewcommand\color[2][]{}%
  }%
  \providecommand\transparent[1]{%
    \errmessage{(Inkscape) Transparency is used (non-zero) for the text in Inkscape, but the package 'transparent.sty' is not loaded}%
    \renewcommand\transparent[1]{}%
  }%
  \providecommand\rotatebox[2]{#2}%
  \newcommand*\fsize{\dimexpr\f@size pt\relax}%
  \newcommand*\lineheight[1]{\fontsize{\fsize}{#1\fsize}\selectfont}%
  \ifx\svgwidth\undefined%
    \setlength{\unitlength}{5173.3629506bp}%
    \ifx\svgscale\undefined%
      \relax%
    \else%
      \setlength{\unitlength}{\unitlength * \real{\svgscale}}%
    \fi%
  \else%
    \setlength{\unitlength}{\svgwidth}%
  \fi%
  \global\let\svgwidth\undefined%
  \global\let\svgscale\undefined%
  \makeatother%
  \begin{picture}(1,0.29421996)%
    \lineheight{1}%
    \setlength\tabcolsep{0pt}%
    \put(0,0){\includegraphics[width=\unitlength,page=1]{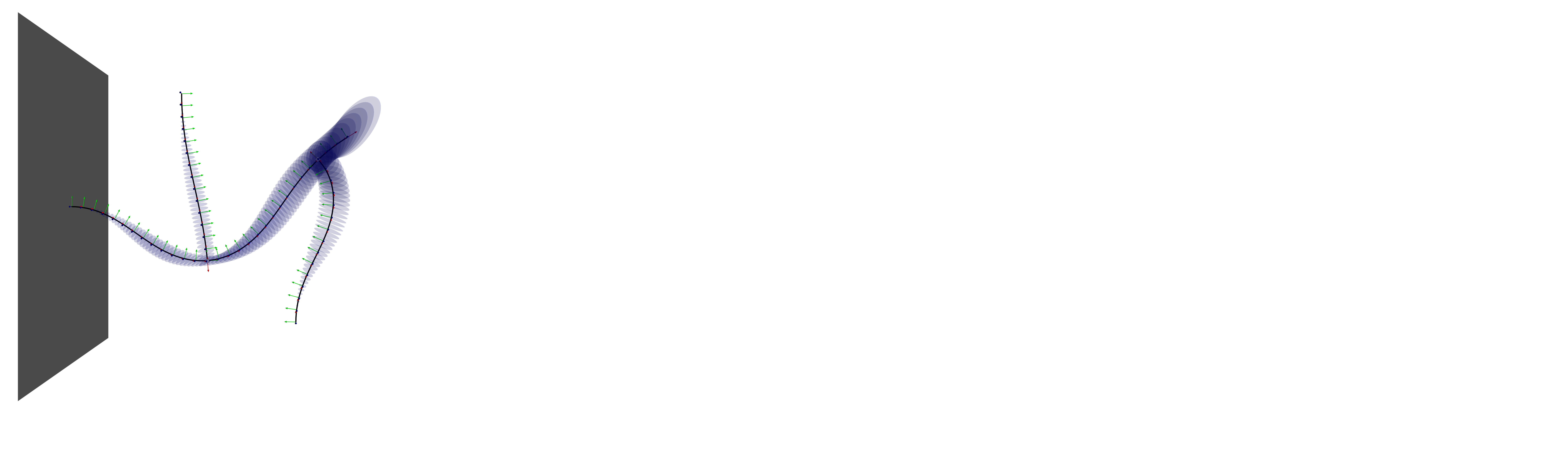}}%
    \put(0.12694233,0.00292893){\makebox(0,0)[t]{\lineheight{1.25}\smash{\begin{tabular}[t]{c}Reconfigurable Parallel Continuum Robot\end{tabular}}}}%
    \put(0,0){\includegraphics[width=\unitlength,page=2]{sota_examples.pdf}}%
    \put(0.37109899,0.00292893){\makebox(0,0)[t]{\lineheight{1.25}\smash{\begin{tabular}[t]{c}Continuum Stewart-Gough Platform\end{tabular}}}}%
    \put(0,0){\includegraphics[width=\unitlength,page=3]{sota_examples.pdf}}%
    \put(0.6018876,0.00292893){\makebox(0,0)[t]{\lineheight{1.25}\smash{\begin{tabular}[t]{c}Continuum Delta Robot\end{tabular}}}}%
    \put(0.86602759,0.00292893){\makebox(0,0)[t]{\lineheight{1.25}\smash{\begin{tabular}[t]{c}Arbitrary Coupled Continuum Rod Topology\end{tabular}}}}%
    \put(0,0){\includegraphics[width=\unitlength,page=4]{sota_examples.pdf}}%
  \end{picture}%
\endgroup%

	\parbox{10cm}{\vspace*{-8cm} \hspace{5.75cm} rigid}
	\caption{Application of the proposed state estimation approach to various system topologies, from left to right: Reconfigurable Parallel Continuum Robot \cite{Anderson2017}, Continuum Stewart-Gough Platform \cite{Black2018}, Continuum Delta Robot \cite{Yang2018}, Arbitrary Coupled Continuum Rod Topology. The most likely system state is shown considering the prior and present coupling constraints without additional measurements. Uncertainties are visualized using blue uncertainty ellipsoids.}
	\label{fig:sota_examples}
\end{figure*}

Fig.~\ref{fig:figure1} shows the posterior estimate for an example system topology.
In this example, two continuum robots are coupled to an end-effector.
The posterior estimate is obtained by considering a noisy measurement of the end-effector pose in addition to our prior formulation and the coupling constraints in the system.
The figure shows the most likely state of the overall system in combination with its uncertainty, visualized with blue uncertainty ellipsoids.

\begin{figure*}[t]
	\centering
	\includegraphics[width=1\linewidth]{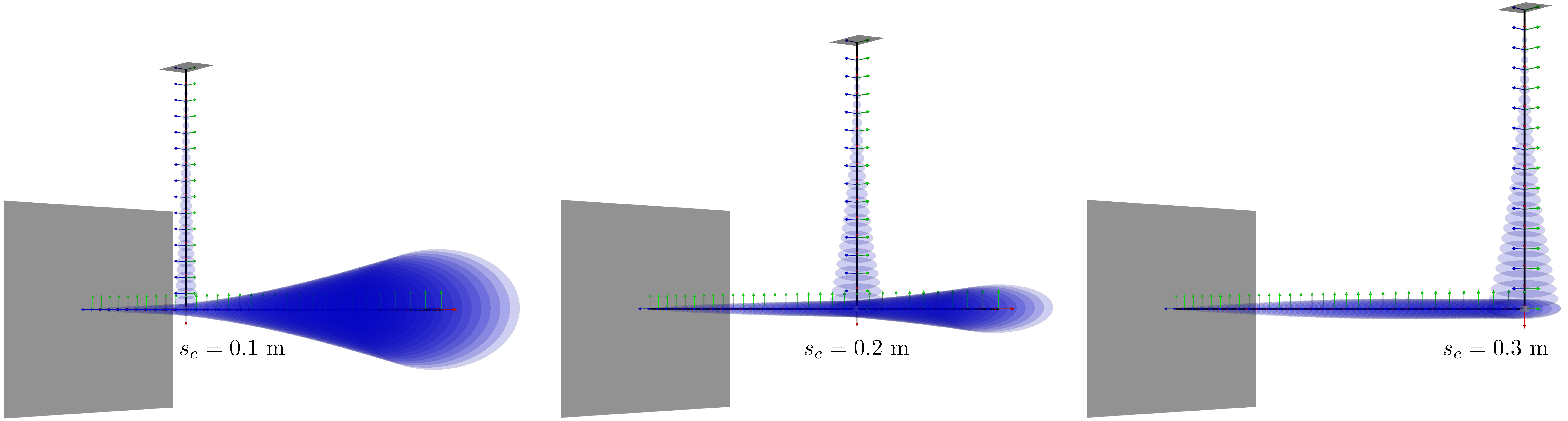}
	\caption{Influence of coupling location $s_c$ on the resulting state estimate of two rigidly coupled continuum robots.}
	\label{fig:coupling_location}
\end{figure*}

\section{Simulations}

We demonstrate our state estimation method's effectiveness through simulations. 
Initially, we apply our framework to various continuum robot topologies to illustrate its versatility. 
Subsequently, we examine the impact of different coupling locations, joint types, and sensor configurations on the accuracy and uncertainty of the state estimate. 
Finally, we assess the framework's computational efficiency.

\subsection{Application to Different Topologies}

Fig.~\ref{fig:sota_examples} shows the application of our proposed state estimation approach to a variety of different system topologies.
In each example, the most likely robot state according to the prior model and coupling constraints without measurements is shown together with 3$\sigma$-uncertainty ellipsoids.
Each state estimation problem converged in less than 10 iterations.

Our first topology features a reconfigurable parallel continuum robot \cite{Anderson2017}, with three robots connected in parallel. 
We then apply our state estimation to a continuum Stewart-Gough platform \cite{Bryson2014, Black2018}, comprising six links coupled to an end-effector with torsional joints. 
Another topology is the continuum Delta robot \cite{Yang2018}, with three chains, each consisting of two continuum structures connected by a rigid link, linked to a common end-effector. 
Finally, we explore an arbitrary topology of coupled continuum rods, demonstrating our method's versatility.
This topology is particularly interesting, since one of the continuum robots is solely coupled to other manipulators at both its proximal and distal end.
It does not feature coupling to the fixed world at any of its ends.
State estimation approaches for such topologies have not been investigated yet and it remains unclear how existing approaches, such as $\cite{Anderson2017}$, could be adapted to handle them.

The results show that coupling constraints significantly impact the state estimation, particularly reducing uncertainty at coupling points and along the constraint dimensions. 
The latter is notably evident in the continuum Stewart-Gough platform and continuum Delta robot examples.

\subsection{Coupling Location and Joint Type}

Fig.~\ref{fig:coupling_location} shows the state estimate for two rigidly coupled continuum robots using only the prior model and the coupling constraint knowledge.
Here, we highlight the impact of the coupling location $s_c$ on the resulting state estimation, similar to what has been shown in \cite{Anderson2017}.
It can be seen that the uncertainty of the estimate decreases at the coupling location.

\begin{figure}[b!]
	\centering
	\footnotesize
	\def\svgwidth{1\linewidth}
\begingroup%
  \makeatletter%
  \providecommand\color[2][]{%
    \errmessage{(Inkscape) Color is used for the text in Inkscape, but the package 'color.sty' is not loaded}%
    \renewcommand\color[2][]{}%
  }%
  \providecommand\transparent[1]{%
    \errmessage{(Inkscape) Transparency is used (non-zero) for the text in Inkscape, but the package 'transparent.sty' is not loaded}%
    \renewcommand\transparent[1]{}%
  }%
  \providecommand\rotatebox[2]{#2}%
  \newcommand*\fsize{\dimexpr\f@size pt\relax}%
  \newcommand*\lineheight[1]{\fontsize{\fsize}{#1\fsize}\selectfont}%
  \ifx\svgwidth\undefined%
    \setlength{\unitlength}{2915.25002307bp}%
    \ifx\svgscale\undefined%
      \relax%
    \else%
      \setlength{\unitlength}{\unitlength * \real{\svgscale}}%
    \fi%
  \else%
    \setlength{\unitlength}{\svgwidth}%
  \fi%
  \global\let\svgwidth\undefined%
  \global\let\svgscale\undefined%
  \makeatother%
  \begin{picture}(1,0.57420299)%
    \lineheight{1}%
    \setlength\tabcolsep{0pt}%
    \put(0,0){\includegraphics[width=\unitlength,page=1]{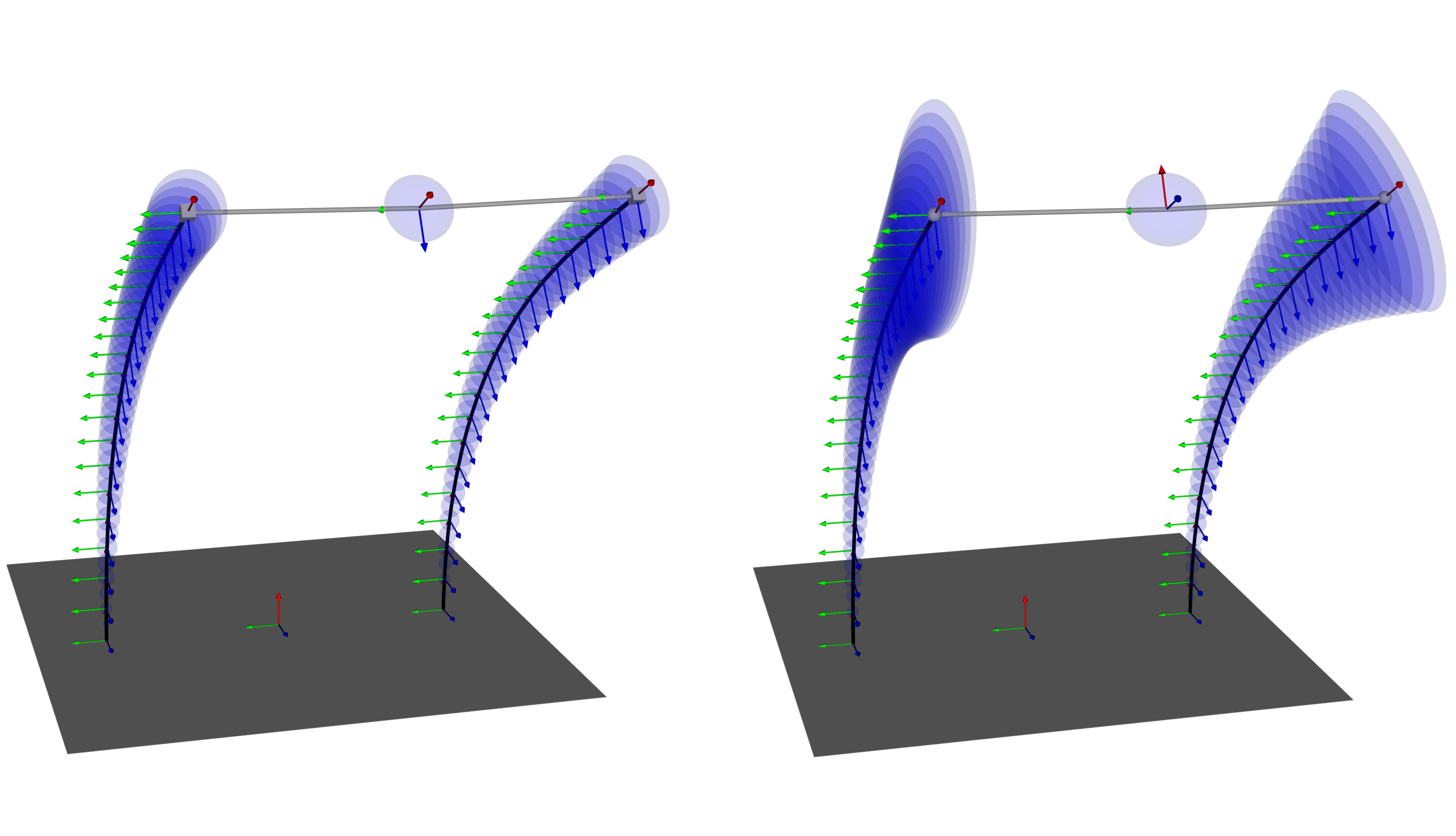}}%
    \put(0.54077309,0.55822022){\color[rgb]{0,0,0}\makebox(0,0)[t]{\lineheight{1.25}\smash{\begin{tabular}[t]{c}End-Effector Position Measurement\end{tabular}}}}%
    \put(0.23282737,0.00462011){\color[rgb]{0,0,0}\makebox(0,0)[t]{\lineheight{1.25}\smash{\begin{tabular}[t]{c}Rigid Joint Coupling\end{tabular}}}}%
    \put(0.75750898,0.00462011){\color[rgb]{0,0,0}\makebox(0,0)[t]{\lineheight{1.25}\smash{\begin{tabular}[t]{c}Spherical Joint Coupling\end{tabular}}}}%
    \put(0,0){\includegraphics[width=\unitlength,page=2]{coupling_joint.pdf}}%
  \end{picture}%
\endgroup%

	\caption{Influence of coupling joint type on the resulting state estimate of two continuum robots coupled to a common end-effector.}
	\label{fig:coupling_joint}
\end{figure}

\begin{table*}\renewcommand{\arraystretch}{1.5}
	\centering
	\caption{Hyperparameters used for quantitative evaluations}
	\label{tab:hyperparameters}
	\begin{tabular}{|c|c|c|c|}
		\hline \multicolumn{2}{|c|}{\textbf{Pose Measurements Covariance}}& \multicolumn{2}{|c|}{\textbf{Strain Measurements Covariance}} \\
		\hline \multicolumn{2}{|c|}{$\mathbf{R}_{m,p} = 2~\mbox{diag}\left(\sigma_p^2~\sigma_p^2~\sigma_p^2~\sigma_o^2~\sigma_o^2~\sigma_o^2\right)$}& \multicolumn{2}{|c|}{$\mathbf{R}_{m,s} = 40~\mbox{diag}\left(\sigma_s^2~\sigma_s^2~\sigma_s^2~\sigma_s^2\right)$} \\		
		\hline $\sigma_p = 2~$mm
		& $\sigma_o = 0.05~$rad  & $\sigma_s = 10~\mu$strain (sim) & $\sigma_s = 100~\mu$strain (exp) \\ \hline 
		\hline \multicolumn{2}{|c|}{\textbf{Coupling Covariance}}& \multicolumn{2}{|c|}{\textbf{Prior Covariance}} \\
		\hline \multicolumn{2}{|c|}{$\mathbf{R}_{c} = 2\cdot10^{-6}~\mbox{diag}\left(1\mathrm{m}^2~1\mathrm{m}^2~1\mathrm{m}^2~1\mathrm{rad}^2~1\mathrm{rad}^2~1\mathrm{rad}^2\right)$ }& \multicolumn{2}{|c|}{$\mathbf{Q}_c = 2~\mbox{diag}\left(0.01\mathrm{m}^2~0.01\mathrm{m}^2~0.01\mathrm{m}^2~1000\mathrm{rad}^2~1000\mathrm{rad}^2~1000\mathrm{rad}^2\right)$} \\	\hline
	\end{tabular}
\end{table*}

\begin{figure*}[b!]
	\centering
	\footnotesize
	\def\svgwidth{1\linewidth}
	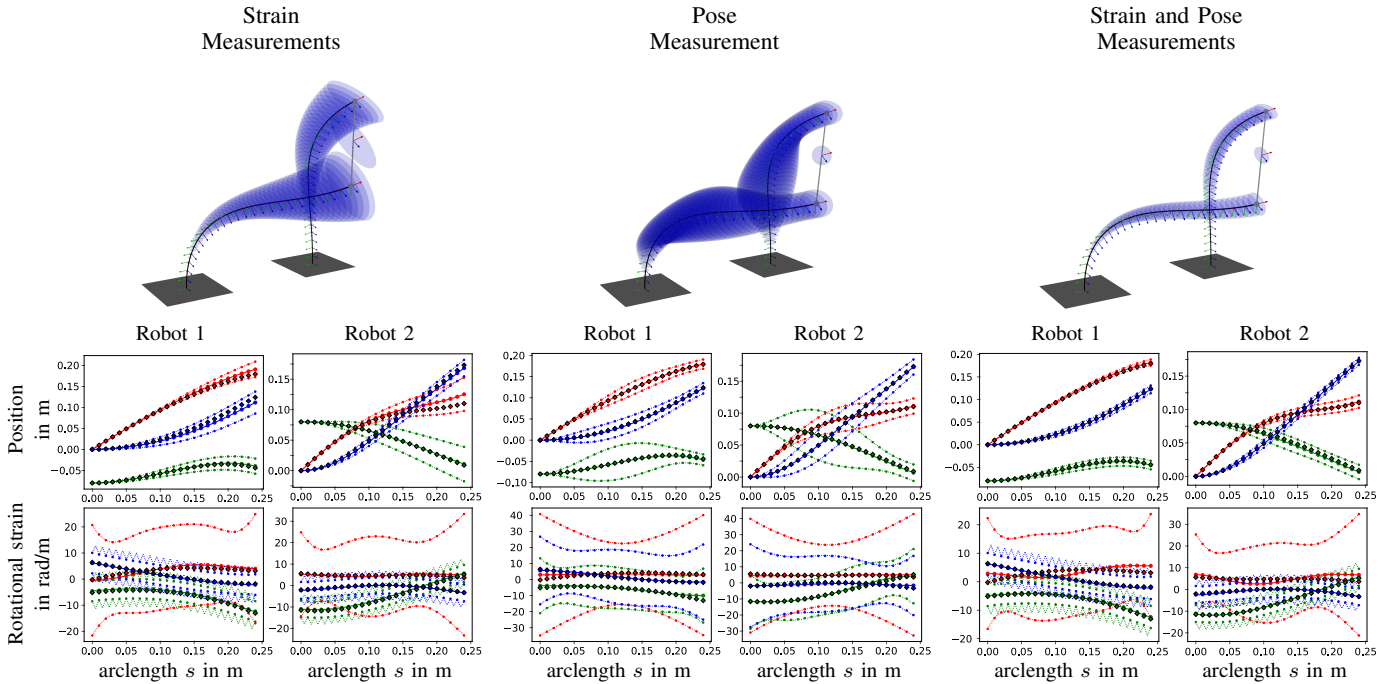
	\caption{State estimates of two continuum robots coupled to a common end-effector using different sensor scenarios in simulation. Left: Using discrete FBG strain measurements along the length of each continuum robot. Middle: Using a pose measurement of the common end-effector. Right: Using both the FBG strain measurements and the pose measurement. For each sensor scenario a rendering of the resulting state estimate is shown including the state mean and 3$\sigma$-uncertainty ellipsoids. Additionally, plots of the continuum robots' position and rotational strain estimates along their arclengths are shown, where the $x$-, $y$- and $z$-components plotted in red, green and blue, respectively. The plots include both the estimated mean, the 3$\sigma$-uncertainty envelopes as well as ground-truth data points depicted as diamond markers.}
	\label{fig:example_configs_sim}
\end{figure*}

Fig.~\ref{fig:coupling_joint} shows the state estimate of two continuum robots coupled to a common end-effector, while considering the prior model, knowledge about coupling constraints and a single end-effector position measurement.
The state estimate is shown for two different coupling joint types.
In the first one, the robots are coupled to the end-effector using rigid joints, while the second one uses spherical joints.
The resulting uncertainty is significantly higher for the spherical joint coupling, as the system is less constrained.
Specifically, since the end-effector orientation is not included in the sensor measurement, rotations in the end-effector pose are possible, leading to higher uncertainties when no rigid connections are present to further constrain its orientation.
This is due to the fact that the continuum robot tips are less constrained in the case of spherical joints and are more likely to deflect, if the end-effector would rotate.
These deflections would particularly happen perpendicular to the direction of the coupling link between continuum robots and end-effector.
Thus, the uncertainty ellipsoids remain small in the directions towards this link as it continues to constrain the continuum robots' tip positions with respect to the end-effector.

\subsection{Sensor Setup Study}

To further showcase the capabilities of the proposed state estimation approach, we investigate its performance using different sensor setups.
For this, we consider two tendon-driven continuum robots (with length $L = 240~\mathrm{mm}$) that are rigidly coupled to a common end-effector (with length $L_\mathrm{ee} = 100~\mathrm{mm}$).
Our to-be-estimated state includes the end-effector pose and $K_n = 25$ discrete states along the length of each robot, leading to a spacing of $\Delta s_{n,k} = 10$ mm.
Throughout the following, we study one particular configuration of this robotic system, in which both robots are bent using their routed tendons, while also applying an external moment to the common end-effector.
The ground-truth data for this configuration is obtained in simulation using the kinetostatic modeling approach for tendon-driven parallel continuum robots presented in \cite{Lilge2022a}.

Using ground-truth data, we simulate noisy FBG sensor measurements and end-effector pose measurements. This involves extracting states from the ground-truth, applying sensor models to calculate expected measurements, and then adding noise drawn from zero-mean normal distributions with standard deviations outlined in Table~\ref{tab:hyperparameters} ($\sigma_p=2$ mm for position, $\sigma_o=0.05$ rad for orientation, and $\sigma_s=10~\mu$strain for FBG strain). Noise addition to FBG strain measurements is straightforward, while pose noise is injected using Lie algebra, similar to the perturbation scheme in \eqref{eq:posepert}.

We then estimate the system state using these noisy measurements and empirically tuned hyperparameters from Table~\ref{tab:hyperparameters}, assuming Kirchhoff rod behaviour by locking translational strains during optimization.
Three sensor scenarios are evaluated: (1) FBG strain measurements at each robot's discrete arclength, (2) a single pose measurement of the common end-effector, and (3) combining FBG strain and pose measurements. 
Results are presented in Fig.~\ref{fig:example_configs_sim}.

Using only FBG strain measurements yields accurate estimates, with errors and uncertainties increasing along the lengths of the continuum robots. 
Notably, uncertainties in bending strains (the $y$- and $z$-components) are much lower than those in twisting strain (the $x$-component), reflecting the FBG sensors' lower sensitivity to twisting deformations.
Additionally, it can again be observed that the uncertainty ellipsoids are smaller in the direction of the existing coupling constraints.

Utilizing solely the end-effector pose measurement results in lower uncertainties near each continuum robot's end and higher uncertainties at unmeasured arclength positions. 
The estimated strains have comparatively high uncertainties due to the lack of direct sensing. 
Nonetheless, the state estimate closely aligns with the ground-truth shapes and strains, suggesting the efficacy of the constant-strain prior in our state estimator.

Lastly, it can be seen that when using both pose and FBG strain measurements the strengths of each sensor type are combined, leading to highly accurate estimates with respect to ground truth with relatively low uncertainties. 
Only the uncertainty of the estimated twisting strain remains high.

\begin{table}[b!]\renewcommand{\arraystretch}{1.5}
	\centering
	\caption{Computation time $t_\mathrm{c}$ and end-effector position error $e_\mathrm{ee}$ for different numbers of estimation nodes $K_n$ and sensor setups}
	\label{tab:comp_time}
	\scriptsize
	\begin{tabular}{|c|c||c|c||c|c|}
		\hline & & \multicolumn{2}{c||}{\textbf{Strain Measurements}} & \multicolumn{2}{c|}{\textbf{Strain \& Pose Measurements}} \\
		\hline \hline $K_n$ & $\Delta s_{n,k}$ & $t_\mathrm{c}$ in ms & $e_\mathrm{ee}$ in mm & $t_\mathrm{c}$ in ms & $e_\mathrm{ee}$ in mm \\
		\hline 25 & 10 mm & 29.36 & 16.01 & 34.15 & 2.96 \\
		13 & 20 mm & 9.24 & 19.80 & 9.03 & 2.72 \\
		7 & 40 mm & 3.70 & 30.96 & 3.85 & 2.79 \\ \hline
	\end{tabular}
\end{table}

\subsection{Computation Time Study}

We evaluate the computational efficiency of our state estimator using the same system topology and configuration as before. 
The computation time depends on the number of discrete states $K_n$ along each continuum robot, which determines the spacing $\Delta s_{n,k}$. 
We test two sensor scenarios: one with FBG strain measurements at each discrete state, and another including an additional end-effector pose measurement. 
The experiments, conducted with three different $K_n$ values and repeated 100 times per scenario, seek to quantify the average computation time $t_c$ and the end-effector position error $e_\mathrm{ee}$.

The results, summarized in Table~\ref{tab:comp_time}, show a significant reduction in computation time for smaller $K_n$, dropping to $3-4$ ms for $K_n=7$. 
However, accuracy at the end-effector decreases when relying solely on FBG and fewer strain measurements.
Inclusion of an end-effector pose measurement compensates for this increased error. 
We conclude that $K_n$ should be chosen as a trade-off between achieved accuracy and computation time depending on the robot topology, available sensor information and application requirements.

\section{Experiments}

\begin{figure}[!b]
	\centering
	\scriptsize
	\def\svgwidth{0.95\linewidth}
\begingroup%
  \makeatletter%
  \providecommand\color[2][]{%
    \errmessage{(Inkscape) Color is used for the text in Inkscape, but the package 'color.sty' is not loaded}%
    \renewcommand\color[2][]{}%
  }%
  \providecommand\transparent[1]{%
    \errmessage{(Inkscape) Transparency is used (non-zero) for the text in Inkscape, but the package 'transparent.sty' is not loaded}%
    \renewcommand\transparent[1]{}%
  }%
  \providecommand\rotatebox[2]{#2}%
  \newcommand*\fsize{\dimexpr\f@size pt\relax}%
  \newcommand*\lineheight[1]{\fontsize{\fsize}{#1\fsize}\selectfont}%
  \ifx\svgwidth\undefined%
    \setlength{\unitlength}{1044.99998078bp}%
    \ifx\svgscale\undefined%
      \relax%
    \else%
      \setlength{\unitlength}{\unitlength * \real{\svgscale}}%
    \fi%
  \else%
    \setlength{\unitlength}{\svgwidth}%
  \fi%
  \global\let\svgwidth\undefined%
  \global\let\svgscale\undefined%
  \makeatother%
  \begin{picture}(1,0.88133971)%
    \lineheight{1}%
    \setlength\tabcolsep{0pt}%
    \put(0,0){\includegraphics[width=\unitlength,page=1]{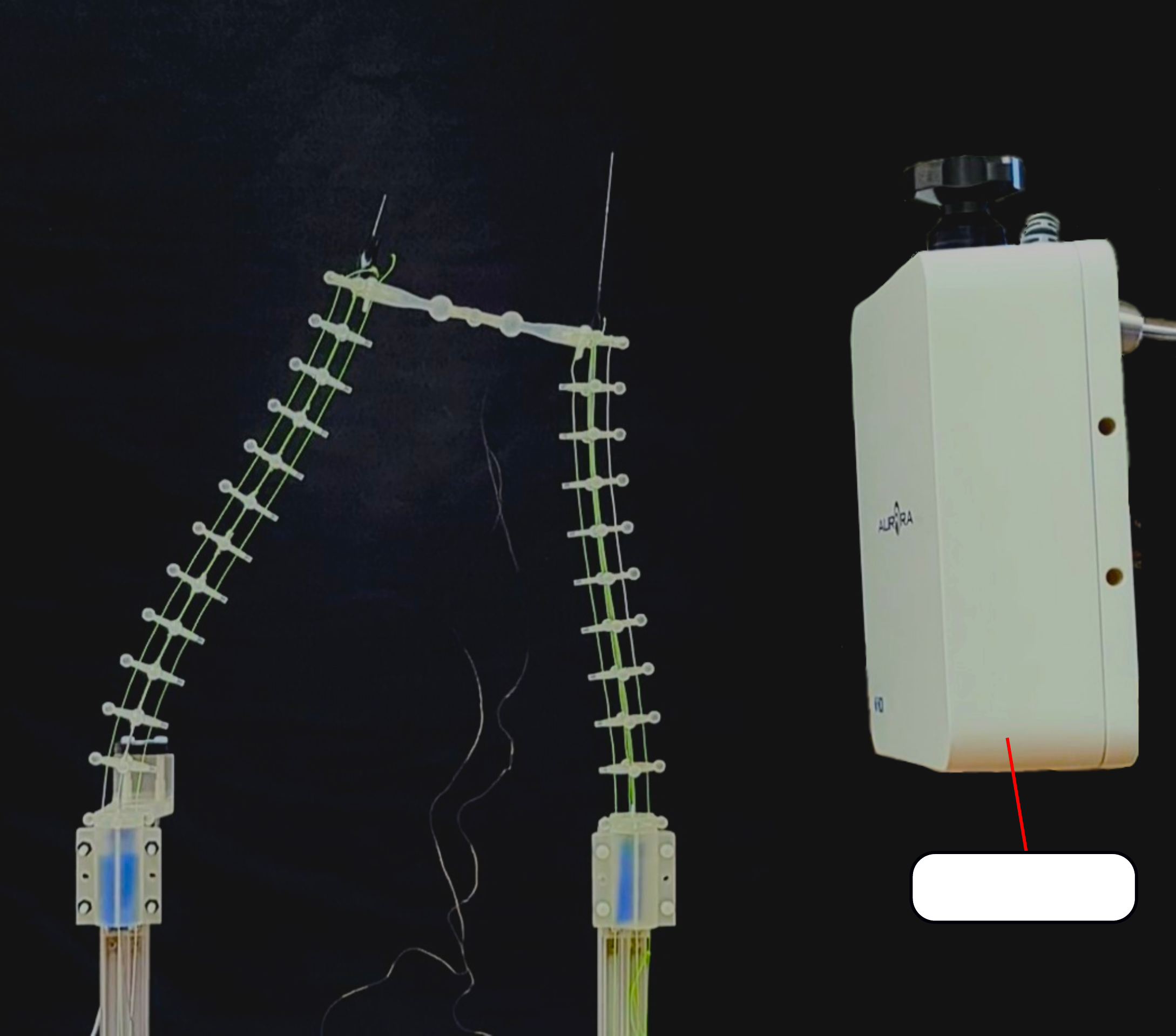}}%
    \put(0.87021721,0.11830863){\color[rgb]{0,0,0}\makebox(0,0)[t]{\lineheight{1.25}\smash{\begin{tabular}[t]{c}Field generator\end{tabular}}}}%
    \put(0,0){\includegraphics[width=\unitlength,page=2]{ex_setup.pdf}}%
    \put(0.34096934,0.11566781){\color[rgb]{0,0,0}\makebox(0,0)[t]{\lineheight{1.25}\smash{\begin{tabular}[t]{c}Reference frame\end{tabular}}}}%
    \put(0,0){\includegraphics[width=\unitlength,page=3]{ex_setup.pdf}}%
    \put(0.37364818,0.82074644){\color[rgb]{0,0,0}\makebox(0,0)[t]{\lineheight{1.25}\smash{\begin{tabular}[t]{c}Electromagnetic\\tracking coil\end{tabular}}}}%
    \put(0,0){\includegraphics[width=\unitlength,page=4]{ex_setup.pdf}}%
    \put(0.34256009,0.3292518){\color[rgb]{0,0,0}\makebox(0,0)[t]{\lineheight{1.25}\smash{\begin{tabular}[t]{c}Fiber Bragg grating\\sensors\end{tabular}}}}%
  \end{picture}%
\endgroup%

	\caption{Continuum robot prototype and experimental setup used for evaluation. The setup includes two  robots rigidly coupled to a common end-effector. An eletromagnetic tracker is attached to this end-effector to obtain pose measurements and both robots feature FBG sensors for strain measurements.}
	\label{fig:exp_setup}
\end{figure}

\begin{figure}[t]
	\centering
	\includegraphics[width=1\linewidth]{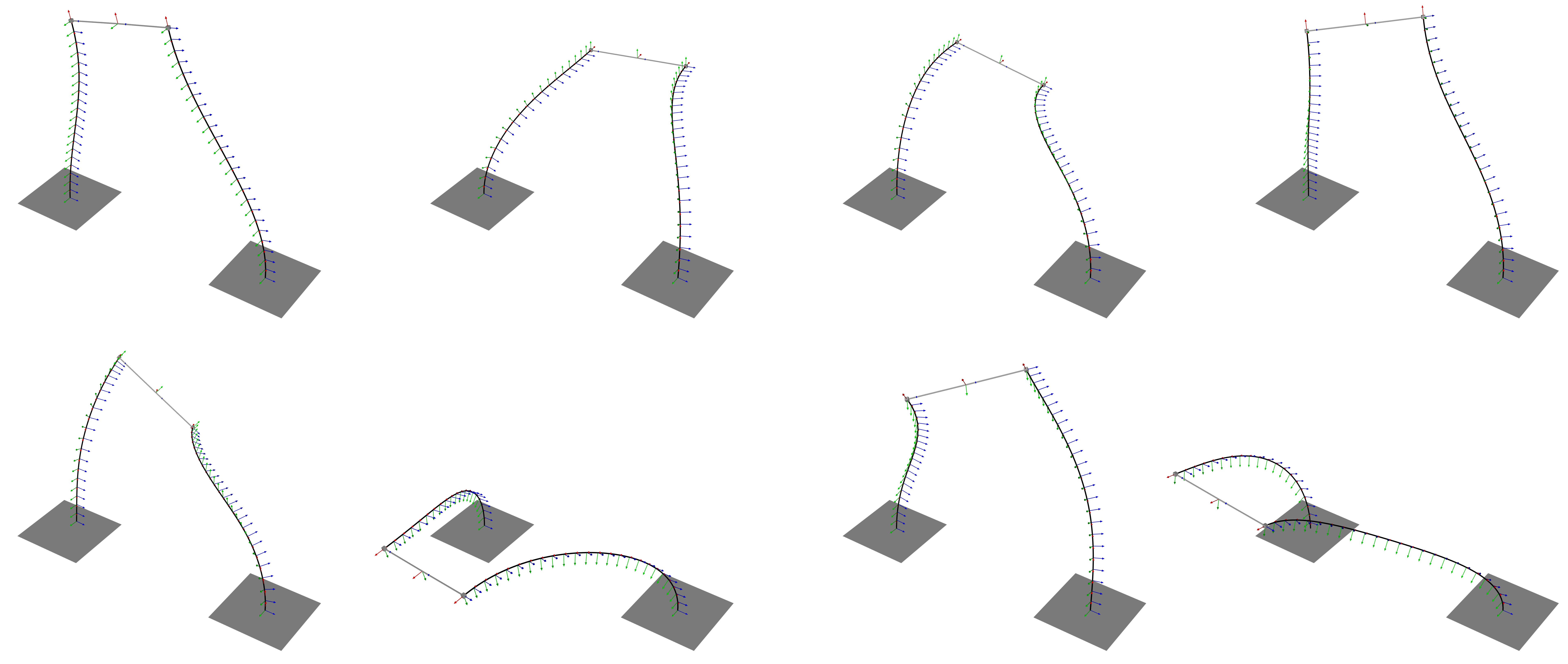}
	\caption{Robot configurations used for experimental validation.}
	\label{fig:exp_data_set}
\end{figure}

Throughout this section, we are evaluating the proposed state estimation approach qualitatively and quantitatively with experiments conducted on a robotic prototype consisting of two coupled continuum robots.

\subsection{Continuum Robot Prototype}

The prototype used for the experimental evaluation is shown in Fig.~\ref{fig:exp_setup}, consisting of two  continuum robots whose tips are rigidly coupled to an end-effector (a bar of length $L_\mathrm{ee} = 100~\mathrm{mm}$).
The continuum robots are of length $L_1 = 240~\mathrm{mm}$ and $L_2 = 200~\mathrm{mm}$, respectively, and feature equally distributed spacer disks at an interval of $20~\mathrm{mm}$.
Each robot is actuated by pulling and releasing four tendons that are routed parallel to their respective backbones and terminate at their distal spacer disks.
The reference frame is defined to coincide with the base frame of the first robot.

An EM tracking coil (Aurora v3, Northern Digital Inc., Canada) is attached at the center of the end-effector to take measurements of its pose.
An additional EM sensor is attached to the base of the first robot to express the pose measurements in the common reference frame.
Additionally, each continuum robot central backbone is equipped with an FBG sensor (MCF-DTG, FBGS, Germany) with gratings spaced with $\Delta s~=~10~\mathrm{mm}$.
The spacer disks and bases of each continuum robot as well as the common end-effector are further equipped with three marker spheres each, used to extract discrete coordinate frames from laser scans.
These frames are used to calibrate and register the different sensor frames with respect to each other, i.e., expressing them all in the common reference frame.
Additionally, they serve as ground-truth measurements when evaluating the accuracy of the proposed state estimation approach.

\subsection{Dataset}

We consider eight configurations of our robotic prototype, demonstrating various regions of its reachable workspace for experimental evaluation (see Fig.~\ref{fig:exp_data_set}). 
These configurations were achieved by manipulating the actuating tendons and applying external forces and moments to the end-effector. 

For each, we recorded the EM tracking coil's pose at the end-effector, FBG sensor strain measurements, and a laser scan of the structure. 
From the laser scans, we extracted discrete frames of robot bases, disks, and the end-effector, all referenced to the first robot's base frame.

\begin{figure*}[t!]
	\centering
	\footnotesize
	\def\svgwidth{1\linewidth}
	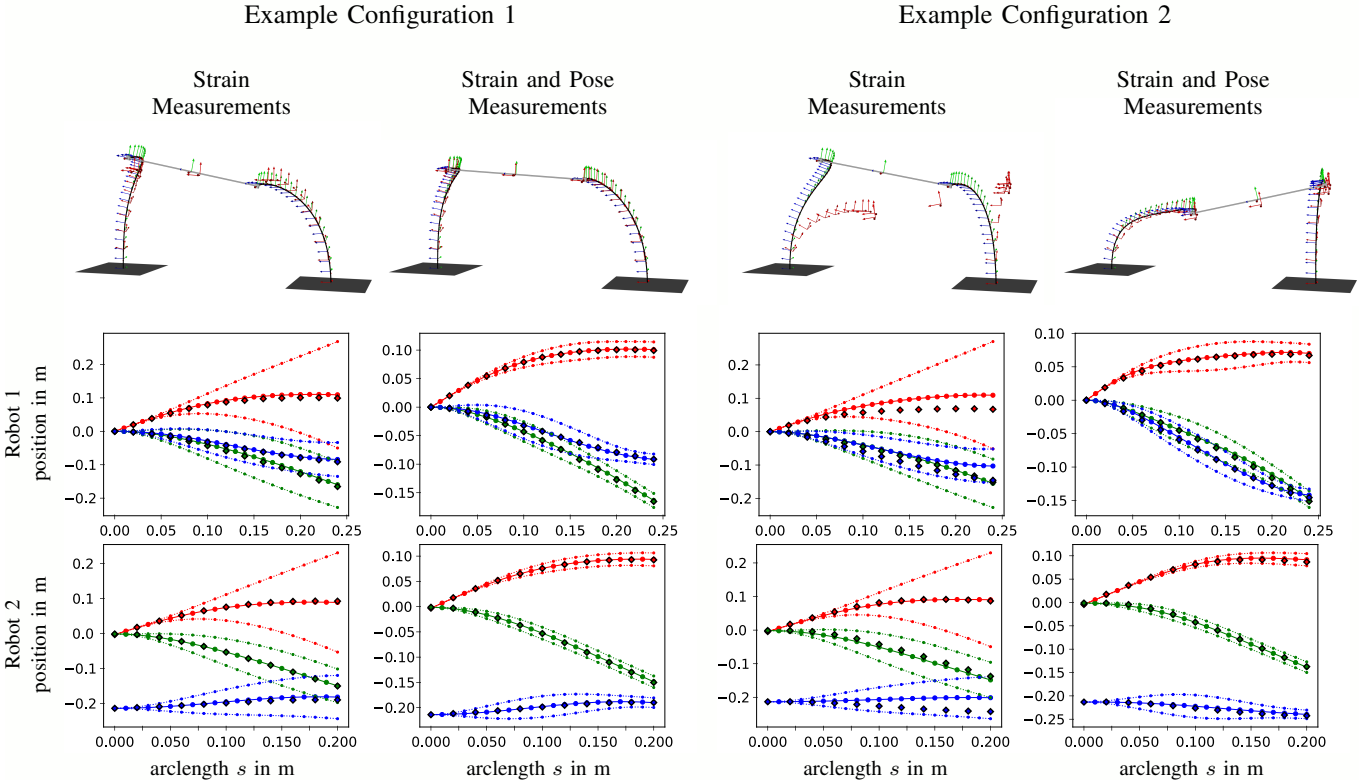
	\caption{State estimates of two continuum robots coupled to a common end-effector using different sensor scenarios for two example experimental configurations. In the first scenario, discrete FBG strain measurements along the length of each continuum robot are used. In the second scenario, an additional measurement of the end-effector pose is used. For each computed state estimate a rendering of the resulting state mean is shown in comparison to the ground-truth states depicted as red frames. Additionally, plots of the continuum robots' position estimates along their arc lengths are shown, where the $x$-, $y$- and $z$-components plotted in red, green and blue, respectively. The plots include both the estimated mean, the 3$\sigma$-uncertainty envelopes as well as ground-truth data points depicted as diamond markers.}
	\label{fig:example_configs_exp}
\end{figure*}

\subsection{Calibration}

In the following, we discuss the calibration of the different sensors utilized in the experimental setup, in order to express their quantities in the common reference frame.

\subsubsection{Fiber Bragg Grating Sensors}

According to the sensor model discussed in Sec.~\ref{sec:sensor_model}, each measurement obtained from the employed Fiber-Bragg-Grating sensors needs to be expressed in the local coordinate frame, i.e., body frame, of the corresponding robot.
For this, the initial orientation of the employed sensing fiber with respect to this local frame of the continuum robot must be known.

In order to determine these orientations, which can each be expressed by a single angle of rotation $\theta_\mathrm{offset}$, a simple calibration routine is performed.
During calibration, each continuum robot is considered individually in an uncoupled state.
The robot is then bent into four different directions, utilizing the routed tendons.
In each bent configuration, we obtain the FBG strain measurements as well as the discrete coordinate frames of each disk expressed in the robot's base frame from laser scans.
Making use of the relationship between the global and local pose variables in \eqref{eq:local_global1}, the discrete frames are then used to approximate the translational and rotational strains along the robot's length.
Using our FBG sensor model from Sec.~\ref{sec:sensor_model}, these strains are then used to obtain the expected FBG measurements given the bending state of the robot.

Using this workflow, we employ an optimization scheme to find the unknown sensor orientation $\theta_\mathrm{offset}$ that minimizes the difference between the expected and obtained FBG strain measurements for both robots.
The resulting angles are $\theta_\mathrm{offset,1}=-14.4^\circ$ for the first robot and $\theta_\mathrm{offset,2}=-41.2^\circ$ for the second.
It is noted that even after calibration relatively high remaining maximum errors between the expected and obtained FBG strain measurement persist.
The remaining maximum errors result in $e_\mathrm{calib,max,1}=473.9~\mu\mathrm{strain}$ for the first robot and $e_\mathrm{calib,max,2}=212.2~\mu\mathrm{strain}$ for the second robot.
This indicates that the utilized FBG sensors might be subject to considerable noise.
This can have several reasons, such as a remaining misalignment of the sensors within the robots' backbones or imperfect temperature compensation.
We will later account for this noise by tuning the corresponding covariances of the FBG strain measurements accordingly.

\subsubsection{Electromagnetic Tracking Sensor}

We aim to calibrate the transformation between the EM tracking sensor at the base of the first robot and the overall reference frame. 
Despite general knowledge of this transformation from system geometry, assembly inaccuracies and parasitic effects may cause errors.
We address this by optimizing for an additional parasitic three-dimensional rotation between the tracking sensor and the reference frame, aiming to minimize position errors between measured and ground-truth end-effector poses. 
To prevent overfitting, we divide the eight configurations into two sets of four. 
The optimization is performed separately on each set, yielding two distinct sets of rotational offsets. 
Each set of offsets is then applied to the configurations not used in its derivation for calibration purposes.
Both optimizations lead to approximately the same rotational offsets, resulting in rotation angles of $\alpha_1=-0.03^\circ$, $\beta_1=0.80^\circ$ and $\gamma_1=4.27^\circ$ for the first set and $\alpha_2=-0.06^\circ$, $\beta_2=1.07^\circ$ and $\gamma_2=4.52^\circ$ for the second.
The remaining average position error between measurements and ground-truth after calibration is $3.29\pm1.37~\mathrm{mm}$.

\subsection{Qualitative Evaluation of Example Configurations}

Using calibrated sensor data, we computed state estimates for recorded robot configurations under two sensor scenarios. 
The first uses FBG strain measurements at each robot's discrete states, while the second adds an end-effector pose measurement. 
The same hyperparameters from simulations (Table~\ref{tab:hyperparameters}) were employed, but due to the high noise in FBG sensors, we increased their measurement covariance by setting $\sigma_s = 100~\mu$strain. 
The state estimation involved $K_1 = 25$ and $K_2 = 21$ discrete states for each robot, with $\Delta s_{n,k} = 10~\mathrm{mm}$.

Fig.~\ref{fig:example_configs_exp} presents state estimates for two configurations from our dataset. 
The first configuration was achieved by tendon actuation, and the second included additional loads on the end-effector, inducing bending and twisting in the robots. 
For the first configuration, both sensor scenarios showed high accuracy compared to ground truth. 
However, in the second configuration, relying solely on FBG strain measurements resulted in notable errors, attributed to sensor noise and low sensitivity to twisting deformations. 
Incorporating the end-effector pose measurement significantly improved accuracy.

\subsection{Quantitative Evaluation}

Table~\ref{tab:ex_results} presents a summary of the experimental validation results using all eight configurations from our dataset. 
It details position and orientation errors at the distal ends of each continuum robot and the common end-effector, for both sensor scenarios: using only FBG strain measurements and including an additional end-effector pose measurement.

\begin{figure}[b!]
	\centering
	\scriptsize
	\def\svgwidth{1\linewidth}
	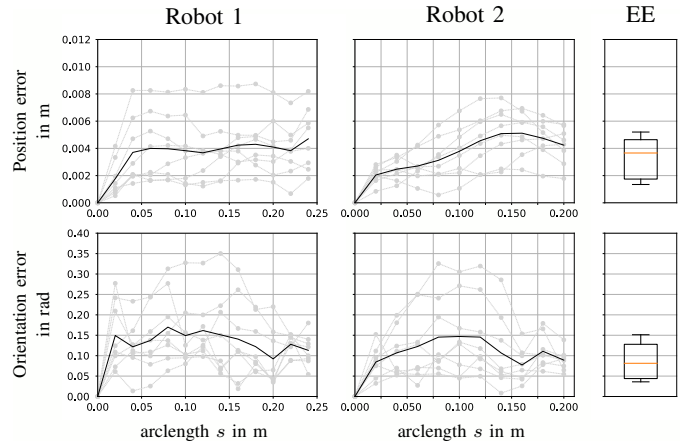
	\caption{Experimental state estimation position and orientation errors using FBG strain measurements and the end-effector pose measurement. The errors along the arclength of each continuum robot are plotted in grey for each configuration, while the black plots show the mean errors. The errors of the end-effector position and orientation are provided as boxplots.}
	\label{fig:error_plots_experiments}
\end{figure}

\begin{table*}\renewcommand{\arraystretch}{1.5}
	\centering
	\caption{State estimation accuracy of experiments for different sensor scenarios}
	\label{tab:ex_results}
	\begin{tabular}{|l||c|c|c||c|c|c||c|c|c||c|c|c|}
		\hline & \multicolumn{6}{c||}{\textbf{Strain Measurements}} & \multicolumn{6}{c|}{\textbf{Strain and Pose Measurements}} \\
		\hline \hline & \multicolumn{3}{c||}{Pos. error in mm} &  \multicolumn{3}{c||}{Rot. error in $^\circ$} & \multicolumn{3}{c||}{Pos. error in mm} &  \multicolumn{3}{c|}{Rot. error in $^\circ$} \\
		\hline \hline & mean & std & max &  mean & std & max & mean & std & max & mean & std & max \\
		\hline Robot 1 Tip & 35.90 & 28.71 & 92.29 & 17.75 & 11.76 & 41.42 & 4.70 &  2.27 & 8.19 & 6.48 & 2.19 & 10.32 \\
		Robot 2 Tip & 29.47 & 14.01 & 44.42 & 18.60 & 10.90 & 40.92 & 4.24 & 1.38 & 5.75 & 5.11 & 1.48 & 7.94 \\
		End-Effector &  29.49 & 20.03 & 63.40 & 19.37 & 11.72 & 42.93 & 3.32 & 1.60 & 5.20 & 5.02 & 2.72 & 8.67 \\ \hline
	\end{tabular}
\end{table*}

The errors are notably higher when relying solely on FBG strain measurements. The mean position errors at the distal ends of the continuum robots are 35.90~mm and 29.47~mm, representing 15.0\% and 19.7\% of their lengths, respectively.
The mean position and orientation errors at the end-effector are 29.49~mm and 19.37$^\circ$.
However, these errors significantly reduce when incorporating an additional end-effector pose measurement. 
In this scenario, the average position errors at the distal ends decrease to  4.70~mm and 4.24~mm, about 2.0\% and 2.1\% of their lengths, respectively.
The average end-effector errors are reduced to 3.32~mm and 5.02$^\circ$.

\begin{figure}[b!]
	\centering
	\footnotesize
	\def\svgwidth{1\linewidth}
	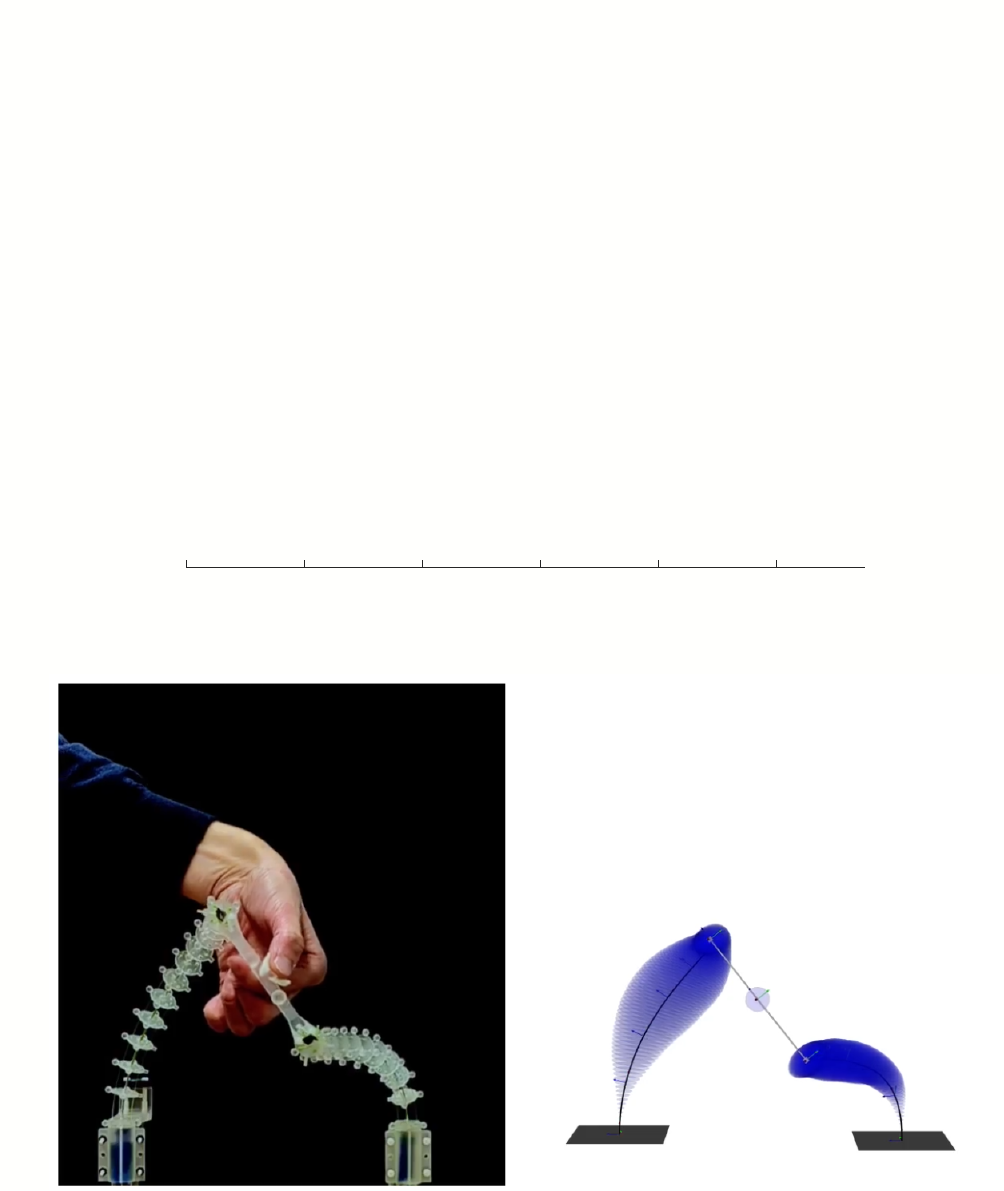
	\caption{Top: Computation times of the proposed state estimation approach for a sequences of configurations of two coupled continuum robots. Bottom: Side-by-side comparison of an example configuration from this sequence and the resulting state estimate.}
	\label{fig:real_time_example}
\end{figure}

Fig.~\ref{fig:error_plots_experiments} provides detailed error analysis for the second sensor scenario. 
It features position and orientation error plots along each continuum robot's arclength, and boxplots illustrating the position and orientation errors at the end-effector.

\subsection{Quasi-Static Real-Time State Estimation}

We experimentally validate the real-time capabilities of our state estimation approach. 
The robotic prototype was actuated into various bending and twisting shapes, while recording the motion sequence with FBG strain and end-effector pose measurements at 40 Hz. 
State estimates were computed at each time step using these measurements, employing the same hyperparameters as before but with reduced discrete states ($K_1 = 7$, $K_2 = 6$) for computational efficiency. 
This results in FBG strain measurements at 40~mm intervals.

Fig.~\ref{fig:real_time_example} displays computation times throughout the motion sequence and compares an example configuration with its state estimate. 
The average computation time is 6.26~ms, with a standard deviation of 1.21~ms and a maximum of 28.23~ms. 
Computation times increase with more complex deformations but mostly remain below the 40~Hz sensor update rate, indicating real-time applicability. 
The majority of computations are even faster than a 100~Hz rate. 
To additionally optimize performance, reducing convergence thresholds or iteration counts could reduce computation times at the expense of accuracy. 
A `warm start' approach, using the previous state estimate as the initial guess, could further enhance efficiency. 
A full side-by-side visualization of the recorded motion sequence and the resulting state estimates over time can be found in the video included in the appendices of this manuscript.

\subsection{Extension of the Continuum Robot Topology}

\begin{figure}[!b]
	\centering
	\scriptsize
	\def\svgwidth{0.825\linewidth}
\begingroup%
  \makeatletter%
  \providecommand\color[2][]{%
    \errmessage{(Inkscape) Color is used for the text in Inkscape, but the package 'color.sty' is not loaded}%
    \renewcommand\color[2][]{}%
  }%
  \providecommand\transparent[1]{%
    \errmessage{(Inkscape) Transparency is used (non-zero) for the text in Inkscape, but the package 'transparent.sty' is not loaded}%
    \renewcommand\transparent[1]{}%
  }%
  \providecommand\rotatebox[2]{#2}%
  \newcommand*\fsize{\dimexpr\f@size pt\relax}%
  \newcommand*\lineheight[1]{\fontsize{\fsize}{#1\fsize}\selectfont}%
  \ifx\svgwidth\undefined%
    \setlength{\unitlength}{3011.99995386bp}%
    \ifx\svgscale\undefined%
      \relax%
    \else%
      \setlength{\unitlength}{\unitlength * \real{\svgscale}}%
    \fi%
  \else%
    \setlength{\unitlength}{\svgwidth}%
  \fi%
  \global\let\svgwidth\undefined%
  \global\let\svgscale\undefined%
  \makeatother%
  \begin{picture}(1,0.88379813)%
    \lineheight{1}%
    \setlength\tabcolsep{0pt}%
    \put(0,0){\includegraphics[width=\unitlength,page=1]{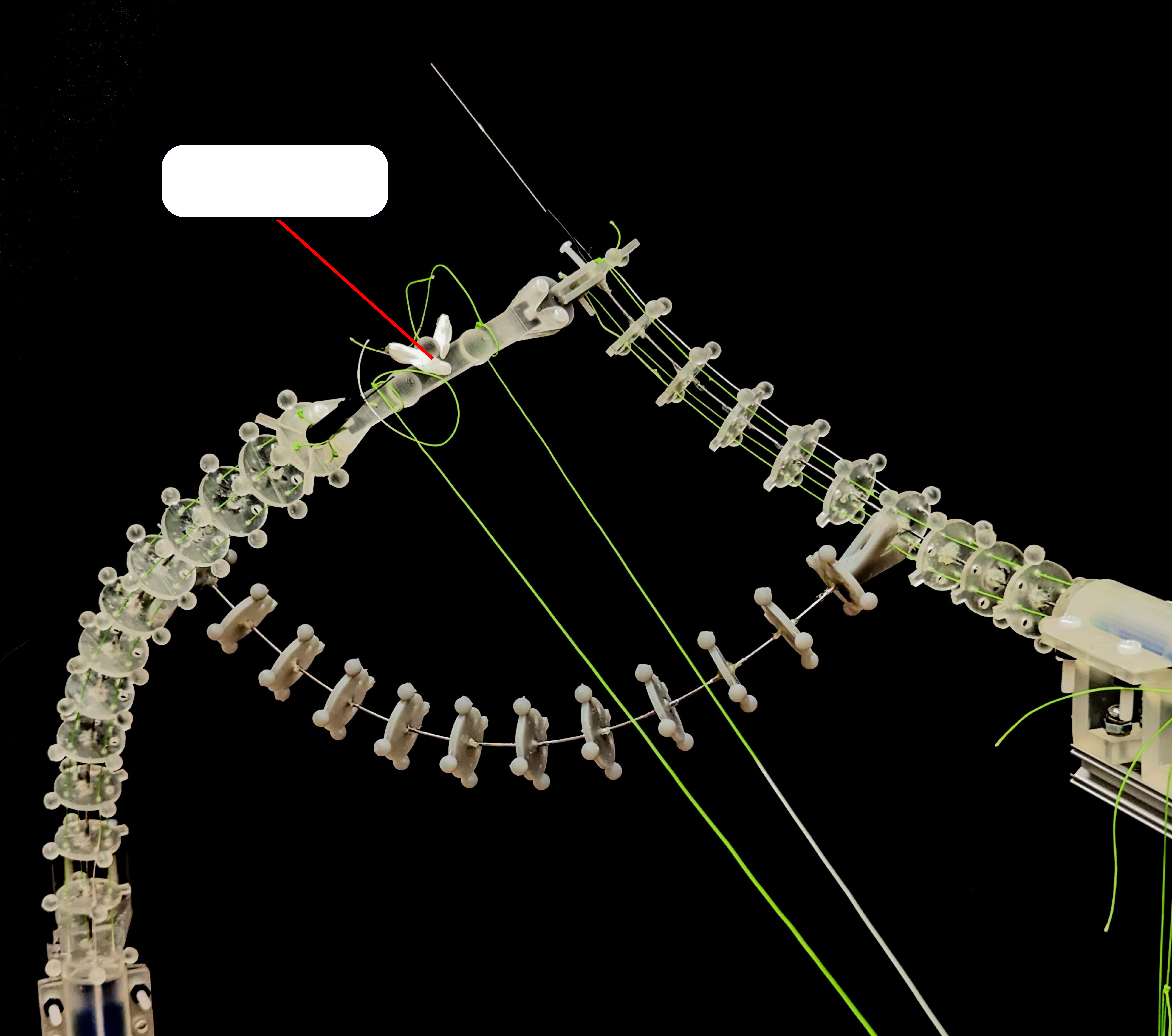}}%
    \put(0.23349713,0.72095183){\color[rgb]{0,0,0}\makebox(0,0)[t]{\lineheight{1.25}\smash{\begin{tabular}[t]{c}Pose Sensor\end{tabular}}}}%
    \put(0,0){\includegraphics[width=\unitlength,page=2]{ex_setup_extension.pdf}}%
    \put(0.27424118,0.11613357){\color[rgb]{0,0,0}\makebox(0,0)[t]{\lineheight{1.25}\smash{\begin{tabular}[t]{c}Robot 1\end{tabular}}}}%
    \put(0,0){\includegraphics[width=\unitlength,page=3]{ex_setup_extension.pdf}}%
    \put(0.83540185,0.61055953){\color[rgb]{0,0,0}\makebox(0,0)[t]{\lineheight{1.25}\smash{\begin{tabular}[t]{c}Robot 2\end{tabular}}}}%
    \put(0,0){\includegraphics[width=\unitlength,page=4]{ex_setup_extension.pdf}}%
    \put(0.58991735,0.09754244){\color[rgb]{0,0,0}\makebox(0,0)[t]{\lineheight{1.25}\smash{\begin{tabular}[t]{c}Robot 3\end{tabular}}}}%
    \put(0,0){\includegraphics[width=\unitlength,page=5]{ex_setup_extension.pdf}}%
  \end{picture}%
\endgroup%

	\caption{Continuum robot topology utilized for additional experimentation. Two continuum robots are rigidly coupled to a common end-effector, which features a electromagnetic tracking coil for pose sensing. A third continuum robot is coupled to both of these manipulators at its proximal and distal ends.}
	\label{fig:exp_setup_extension}
\end{figure}

\begin{figure*}[t!]
	\centering
	\scriptsize
	\def\svgwidth{0.9\linewidth}
	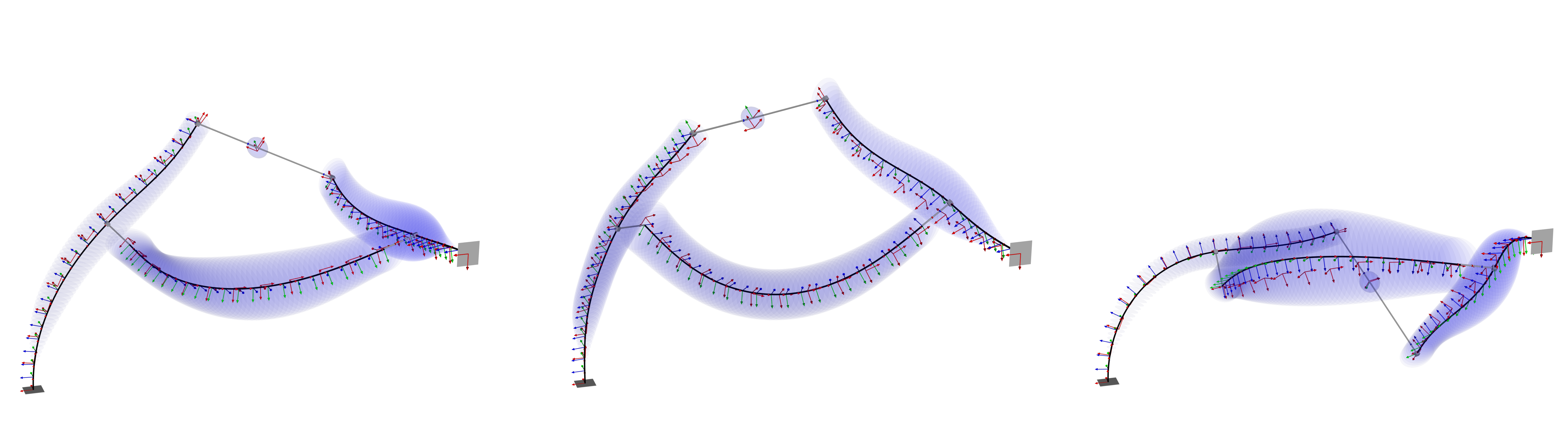
	\caption{State estimates of the extended continuum robot topology in three experimental configurations. For each computed state estimate a rendering of the resulting state mean is shown in comparison to the ground-truth states depicted as red frames. Uncertainty ellipsoids are shown in blue.}
	\label{fig:render_revisions}
\end{figure*}

\begin{figure}[b!]
	\centering
	\scriptsize
	\def\svgwidth{1\linewidth}
	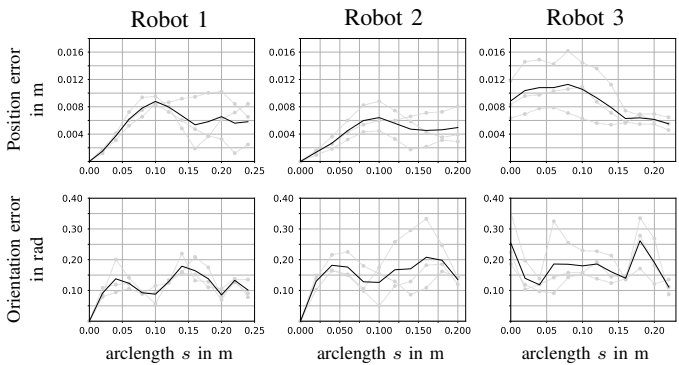
	\caption{Experimental state estimation position and orientation errors for the extended continuum robot topology. The errors along the arclength of each continuum robot are plotted in grey for each configuration, while the black plots show the mean errors.}
	\label{fig:error_plots_revisions}
\end{figure}

To further evaluate the capabilities of the proposed state estimation approach, an additional experiment featuring a more intricate topology of coupled continuum robots is conducted.
The new topology features three continuum robots as depicted in Fig.~\ref{fig:exp_setup_extension}.
The first two continuum robots, which are the same ones used in the previous experiment, are coupled to a common end-effector using rigid connections.
This time, these two robots are organized in a more spatial arrangement, where one robot is orientated vertically and the other horizontally.
An additional third robot of length $L_3 = 220$ mm is considered in this topology, which is coupled to both the first and second robot at its proximal and distal ends, again using rigid connections.
For state estimation, a single noisy pose measurement at the end-effector is considered (see Fig.~\ref{fig:exp_setup_extension}).
In total, three different configurations of this topology are evaluated, which are achieved by actuating the tendons of the first and second robot as well as applying external loads to the end-effector for additional deflections.

For state estimation, the same hyperparameters as before are used, with some exceptions.
First, we increase the prior covariance of the third robot to $\mathbf{Q}_{c,3} = 20\cdot\mathbf{Q}_{c}$.
This effectively means that we assume that this robot is more likely to follow constant-strain deformations compared to the first two robots.
Second, we observe that the rigid coupling between the third and second robot is not perfect and some twisting rotations of the third robot at this location are possible.
This is a result from the manual manufacturing process for the prototype, which relies on glue to keep the disks attached to the backbones.
However, we can account for this by increasing the uncertainty of the coupling constraint $\mathbf{R}_{c,3}$ for the third robot in this degree of freedom (rotation around the $x$-axis) from $2\cdot10^{-6}~\mathrm{rad}^2$ to $2\cdot10^{-1}~\mathrm{rad}^2$.
Lastly, We enforce nominal strain boundary conditions to every continuum robot end that is not attached to the fixed world, which seems to yield the best performance in terms of accuracy.
During state estimation, we consider $K_1 = 25$, $K_2 = 21$ and $K_3 = 23$ discrete estimation nodes for the three robots.

The state estimation results for the three considered configurations, utilizing the single pose measurement of the end-effector in each case, are depicted in Fig.~\ref{fig:render_revisions}.
It includes both the mean and uncertainty of the state estimates as well as ground truth depicted using red frames.
Quantitative position and orientation errors over the arclength of each continuum robot are shown in Fig.~\ref{fig:error_plots_revisions}.
The resulting end-effector accuracies for the three configurations are 3.9 mm and 0.9$^\circ$, 6.4 mm and 3.4$^\circ$, as well as 5.8 mm and 6.7$^\circ$, respectively.
Without further tuning, solving the state estimation problems for this system takes around 80 ms on average, allowing for update rates higher than 10 Hz.
As outlined above, this time can further be minimized by considering less discrete states for the continuum robots or `warm-starting' the optimization with a better initial guess.
Overall, it can be seen that given only a single pose measurement, the state of the continuum robot topology can be accurately determined.
The first and second robot exhibit slightly reduced errors compared to the third robot, as they are directly coupled to the end-effector, whose pose is sensed.

As discussed before, the state estimation for intricate topologies of coupled continuum robots such as the one shown here has not been shown yet.
While the shooting method approach in \cite{Anderson2017} could potentially be adapted to handle topologies beyond those presented in their work, it is unclear how this adaptation would be achieved or how well the method would perform in such cases.
Therefore, demonstrating the ability to handle these complex topologies with our approach highlights its potential as a more general framework.

\section{Conclusion}

This manuscript introduces a novel state estimation approach for systems comprising multiple coupled continuum robots utilizing a sparse Gaussian process regression.
By making use of a Cosserat rod model in combination with a prior favouring constant-strain configurations, the approach can be applied to any continuum robot type and coupling topology.
This makes it particularly useful for state estimation of both parallel and collaborating continuum robots subject to coupling constraints.
Results indicate that accurate state estimates can be achieved, resulting in average end-effector position and orientation errors of 3.32~mm and 5.02$^\circ$ during experiments.
At the same time, fast computation times with average update rates of more than 100~Hz can be realized, making the approach suitable for real-time applications, such as closed-loop control.

Nevertheless, there are a few remaining limitations of our approach.
First, while the assumption of constant strain of our prior model seems sufficient and effective judging by the achieved accuracies, potential knowledge about forces and moments acting on the continuum robots, e.g., from actuation, are not taken into account.
This could potentially lead to limited estimation accuracies in scenarios, in which only little sensor data is available.
Thus, future work could focus on deriving more informed prior models, while aiming to maintain the efficiency of the resulting state estimation approach.
Second, no temporal information is taken into account in the proposed state estimator, as it operates in a quasi-static fashion.
Future work should investigate possible extensions of the proposed method to work in two dimensions, which would allow us to estimate the system states with respect to both spaces, i.e., robot arclength, and time.
Ideally, such an approach would allow us to estimate the continuum robot state continuously and smoothly over time, allowing us to query it at times that do not necessarily correspond to measurement times.
This would further allow us to consider asynchronous measurements of different parts of the robot state in a straightforward manner, i.e., when strain and pose measurements are obtained at different rates.
Lastly, the optimization problem of our state estimation is currently solved using a local Gauss-Newton approach, which might be prone to local minima.
Alternatively, future work could exploit methods that are able to certify and find globally optimal solutions to the estimation problems, such as the one proposed in \cite{Barfoot2023}.

\section*{Acknowledgements}

The authors would like to thank Chloe Pogue for her contributions to designing and assembling the robotic prototype used for the experimental validations throughout this work.

\appendices
\section{Multimedia Appendix}

The appendix includes a video demonstrating the real-time capabilities of the state estimator on the robotic prototype.

\section{Open Source Code}

The C++ code developed for this work is made openly available to the community and can be accessed via \url{https://github.com/SvenLilge/Continuum-MultiRobot-Estimation}.

\bibliographystyle{IEEEtran}
\bibliography{IEEEabrv,refs}

\end{document}